\documentclass[journal]{IEEEtran}
\usepackage{amsmath,amsfonts}
\usepackage{algorithmic}
\usepackage{algorithm}
\usepackage{array}
\usepackage{textcomp}
\usepackage{stfloats}
\usepackage{url}
\usepackage{verbatim}
\usepackage{graphicx}
\usepackage{cite}
\usepackage{color}
\usepackage{booktabs}
\usepackage{threeparttable}
\usepackage{multirow}
\usepackage{makecell}
\usepackage{float}
\usepackage{siunitx}
\usepackage{xurl}
\usepackage[table]{xcolor}
\usepackage{bm}

\newcommand{\etal}{\textit{et al.}}

\usepackage{orcidlink}
\hypersetup{hidelinks}

\makeatletter
\let\MYcaption\@makecaption
\makeatother
\usepackage[font=scriptsize]{subcaption}
\makeatletter
\let\@makecaption\MYcaption
\makeatother

\begin{document}

\title{Joint Identity Verification and Pose Alignment for Partial Fingerprints}

\author{Xiongjun Guan$^{\orcidlink{0000-0001-8887-3735}}$, 
	Zhiyu Pan$^{\orcidlink{0009-0000-6721-4482}}$,
	Jianjiang Feng$^{\orcidlink{0000-0003-4971-6707}}$, ~\IEEEmembership{Member, IEEE}, 
	and Jie Zhou$^{\orcidlink{0000-0001-7701-234X}}$, ~\IEEEmembership{Senior Member, IEEE}
	
	\thanks{
		This work was supported in part by the National Natural Science Foundation of China under Grant 62376132 and 62321005. (\emph{Corresponding author: Jianjiang Feng}.)}
	\IEEEcompsocitemizethanks{
    \IEEEcompsocthanksitem
	The authors are with Department of Automation, Tsinghua University, Beijing 100084, China (e-mail: \url{gxj21@mails.tsinghua.edu.cn}; \url{pzy20@mails.tsinghua.edu.cn}; \url{jfeng@tsinghua.edu.cn}; \url{jzhou@tsinghua.edu.cn}).
	}
	
}

\markboth{Journal of \LaTeX\ Class Files,~Vol.~14, No.~8, August~2021}%
{Guan \MakeLowercase{\textit{et al.}}: Joint Identity Verification and Pose Alignment for Partial Fingerprints}


\maketitle

\begin{abstract}
Currently, portable electronic devices are becoming more and more popular. 
For lightweight considerations, their fingerprint recognition modules usually use limited-size sensors.
However, partial fingerprints have few matchable features, especially when there are differences in finger pressing posture or image quality, which makes partial fingerprint verification challenging.
Most existing methods regard fingerprint position rectification and identity verification as independent tasks, ignoring the coupling relationship between them --- relative pose estimation typically relies on paired features as anchors, and authentication accuracy tends to improve with more precise pose alignment.
In this paper, we propose a novel framework for joint identity verification and pose alignment of partial fingerprint pairs, aiming to leverage their inherent correlation to improve each other.
To achieve this, we present a multi-task CNN (Convolutional Neural Network)-Transformer hybrid network, and design a pre-training task to enhance the feature extraction capability.
Experiments on multiple public datasets (NIST SD14, FVC2002 DB1\_A \& DB3\_A, FVC2004 DB1\_A \& DB2\_A, FVC2006 DB1\_A) and an in-house dataset demonstrate that our method achieves state-of-the-art performance in both partial fingerprint verification and relative pose estimation, while being more efficient than previous methods.
{
Code is available at: https://github.com/XiongjunGuan/JIPNet.
}

\end{abstract}


\section{Introduction}
\IEEEPARstart{F}{ingerprints} are unique patterns composed of ridges and valleys on the finger surface.
Due to its easy collection, high stability and strong recognizability, this biological characteristic exhibits significant application value.
As early as the nineteenth century, researchers had conducted systematic studies on fingerprints \cite{galton1892finger,lee2012lee}.
The reliability of fingerprint recognition gained official recognition in the twentieth century and found extensive application within the judicial sphere \cite{lee2012lee}.
With the advancement of sensors and algorithms, fingerprint recognition technology has been swiftly applied in civilian and commercial fields, such as portable devices, financial services, and access control systems, providing people with convenience and security \cite{maltoni2022handbook}.

\begin{figure}[!t]
	\centering
	\subfloat[\textbf{Sparse key points distribution}]{%
		\includegraphics[height=.25\linewidth]{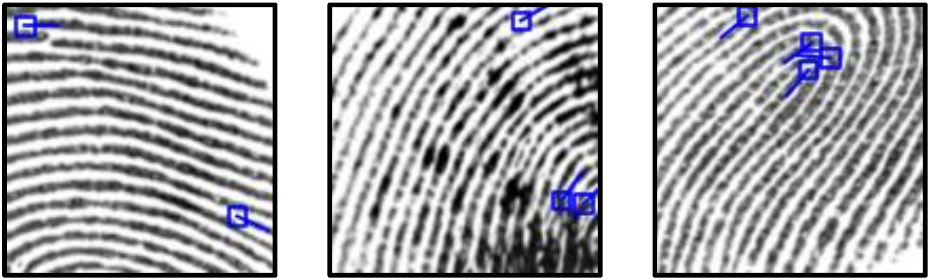}%
		\vspace{-1mm}
	}\hfil
	\par\vspace{2mm}
	\subfloat[\textbf{Similar local texture}]{%
		\includegraphics[height=.25\linewidth]{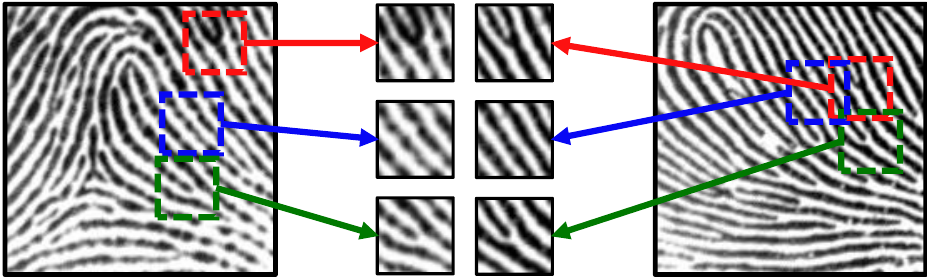}%
		\vspace{-1mm}
	}\hfil
	\par\vspace{2mm}
	\subfloat[\textbf{Significant impression differences}]{%
		\includegraphics[height=.25\linewidth]{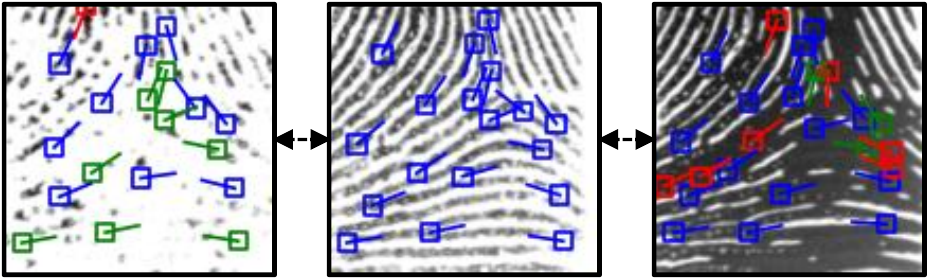}%
		\vspace{-1mm}
	}\hfil
	\caption{Partial fingerprint matching faces three challenges: sparse minutiae distribution, similar local texture and significant modal difference. This figure shows corresponding representative examples: (a) Three weakly-textured fingerprints that lack appropriate landmark feature points. (b) Local ridge patterns of fingerprints from different fingers have high similarity. (c) Fingerprint features in different skin conditions (in this case from left to right are dry, normal and wet respectively) may be missed (green) or incorrectly extracted (red).  Visualized feature points are minutiae extracted by VeriFinger \cite{VeriFinger}.}
	\label{fig:intro_ex}
\end{figure}

Fueled by consumer demand and technological progress, the requirements towards integration, miniaturization, and portability has become prominent within the consumer electronics sector.
This trend compels manufacturers to consistently reduce the size of sensors to accommodate progressively compact  device architectures. 
Fingerprint sensors, currently one of the most popular biometric modules, have also introduced various miniaturized solutions for various applications \cite{maltoni2022handbook}.
However, the size of fingerprint sensors is inevitably limited, leading to a significant reduction in the information available for matching \cite{watson2005effect, fernandez2016small, mathur2016methodology}.
Fig. \ref{fig:intro_ex} shows three representative challenging scenarios in partial fingerprint recognition, which can be even more difficult on fingers with large posture differences or poor quality (such as wear and wrinkles), resulting in a sharp decline in performance of feature extraction and matching \cite{he2022pfvnet}.
This not only reduces the user experience, but also poses great security risks \cite{roy2017masterprint}.

\begin{figure*}[!t]
	\centering
	\captionsetup[subfloat]{justification=centering}
	\subfloat[\textbf{Key points based\cite{cappelli2010minutia}} \label{fig:intro_method_a}]{\includegraphics[height=.38\linewidth]{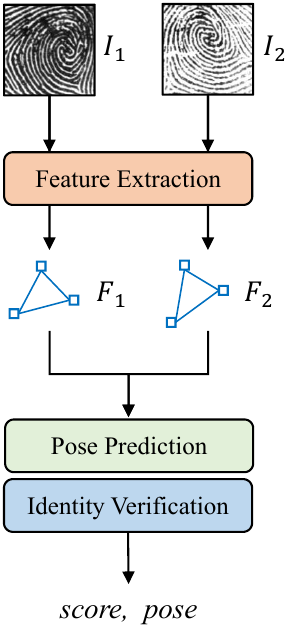}%
	}
	\hfil
	\subfloat[\textbf{Fixed-length region representation based \cite{engelsma2021learning}}\label{fig:intro_method_b}]{\includegraphics[height=.38\linewidth]{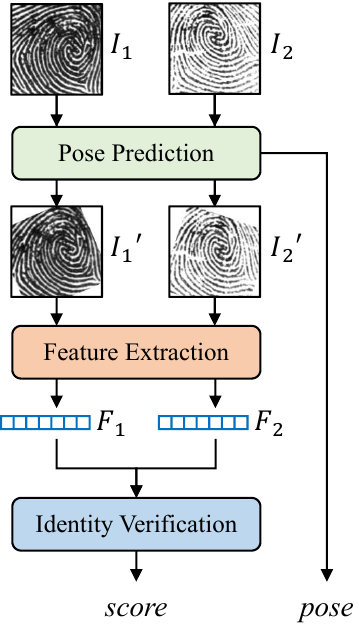}%
	}
	\hfil
	\subfloat[\textbf{Pairwise interrelated feature based\cite{he2022pfvnet}}\label{fig:intro_method_c}]{\includegraphics[height=.38\linewidth]{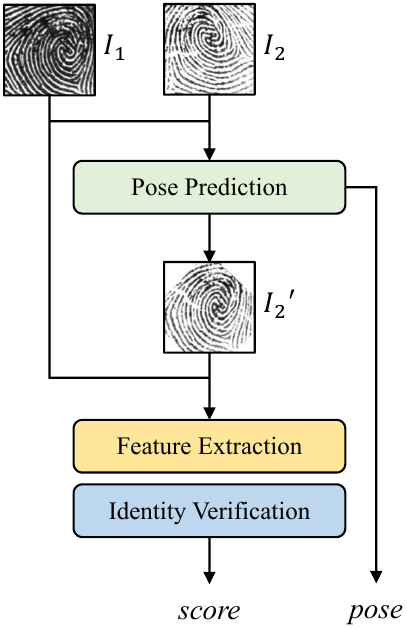}%
	}
	\hfil
	\subfloat[\textbf{Joint estimation based (ours)}\label{fig:intro_method_d}]{\includegraphics[height=.38\linewidth]{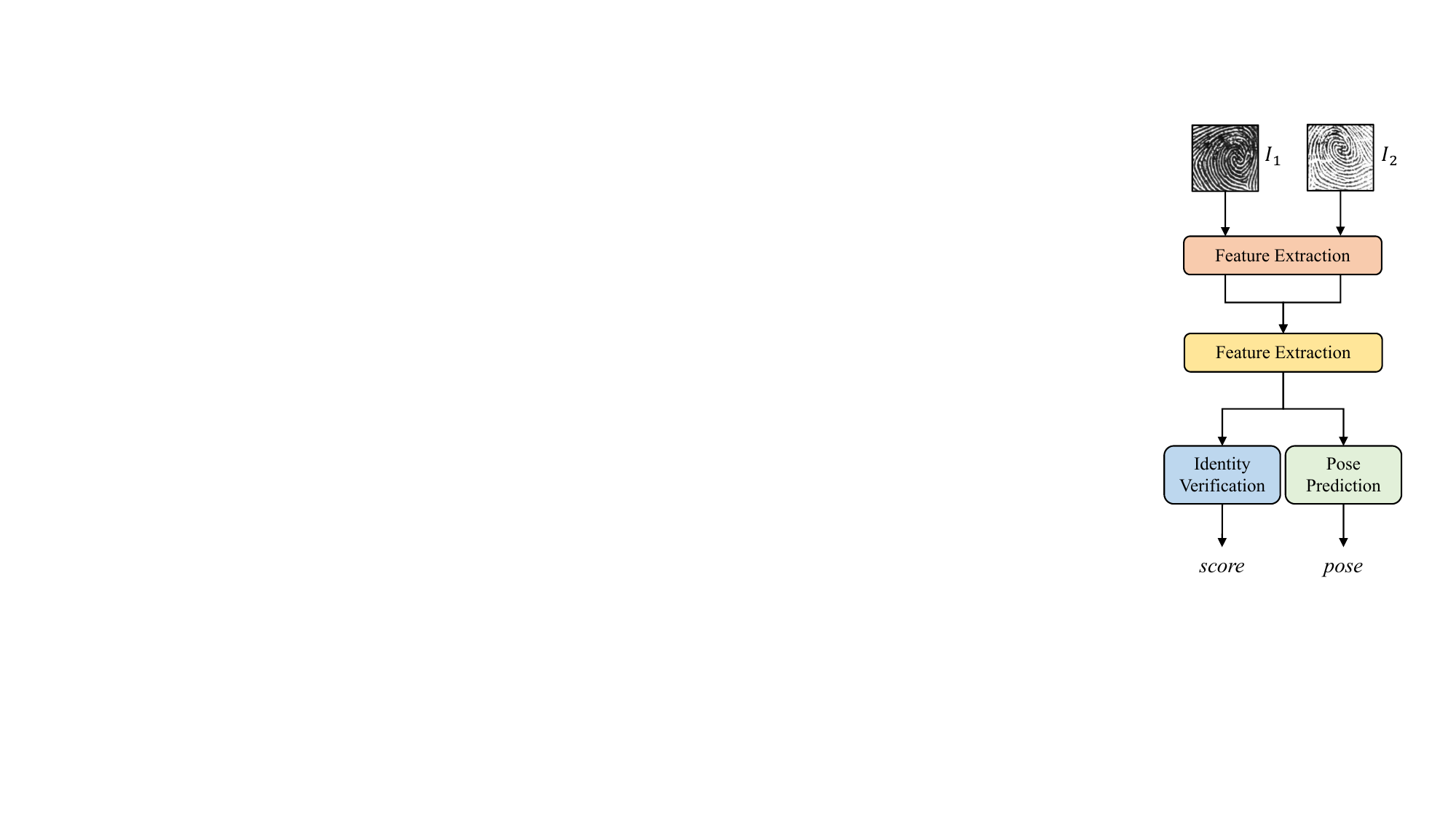}%
	}

	\caption{Frameworks of different fingerprint matching algorithms. The process passed by parallel arrows indicates that the corresponding modules share weights and their functions can be executed independently. \emph{Feature Extraction} in red and gold extract independent and interrelated features from paired data respectively. \emph{Pose Predictor} in (a)(c)(d) and (b) estimate relative and absolute pose respectively.}
	\label{fig:intro_method}
\end{figure*}

A natural idea for partial fingerprint matching is to transfer existing solutions for rolled or plain fingerprint recognition. 
Nevertheless, conventional algorithms usually rely on a sufficient number of key points to establish and compare spatial structural relationships \cite{cappelli2010minutia,choi2011fingerprint}.
While certain studies use neural networks to extract fixed-length region descriptors \cite{engelsma2021learning,gu2021latent, wu2022minutiae, grosz2024afr}, consequently reducing the dependence on point features, these approaches require either large effective area or exceedingly precise alignment.
Considering this, some researchers simultaneously input fingerprint images and extract interrelated features to fully utilize complete information \cite{he2022pfvnet,chen2022query2set}, which effectively improves the matching performance on partial fingerprints.
However, to the best of our knowledge, almost all existing fingerprint matching methods (compared in Fig. \ref{fig:intro_method}) regard pose rectification and identity verification as independent tasks, ignoring the coupling relationship between them.
Specifically, the estimation of relative pose typically relies on paired regional features as anchors, and the accuracy of authentication generally increases with enhanced precision in pose alignment.
{
This is similar to a pictorial jigsaw puzzle where decision-makers need to consider whether (are they neighboring image fragments?\,) and how (what's the reassemble solution?\,) any two pieces fit together \cite{paumard2020deepzzle,markaki2023jigsaw}.
It can be considered as a mixed process, as the relative alignment relationship is confirmed (one of the eight lateral positions for a regular puzzle) simultaneously when two pieces are determined to be adjacent.
On the other hand, an input piece can be classified as outside (mismatched) when it cannot be properly connected to the reference piece at any position.
The comprehensive consideration of identity verification and pose alignment can be transferred to fingerprints, which is feasible because previous works  \cite{he2022pfvnet,duan2023estimating} have demonstrated the complementarity and compatibility between above two tasks.
In addition, the paradigm of joint decision can better balance the robustness and speed of overall process, rather than performing them individually.}

In this paper, we propose a multi-task CNN-Transformer hybrid network which \textbf{J}ointly performs \textbf{I}dentity verification and \textbf{P}ose alignment for partial fingerprints, called \textbf{JIPNet}.
Paired images are input instead of their simplified features to provide more complete information, enabling more flexible and precise analysis.
The intrinsic connection between feature correspondence and position correlation is exploited, aiming to promote each other to achieve higher performance.
{
Unlike previous independent two-stage approaches, the estimation of identity and pose are jointly executed in our proposed framework to leverage their complementarity.
}
Naturally, the efficiency is also improved because these two tasks are compactly integrated.
Inspired by works on natural images \cite{carion2020end,jiang2021cotr, shen2023semi,han2023survey}, we utilize a shared weight convolution layers to extract regional features and transformers to capture both local and global correlations.
Besides, a fingerprint enhancement pre-training task is specially designed to enhance the compatibility with different image modalities and various texture patterns.

Extensive experiments are conducted on multiple datasets, including several public datasets (\emph{NIST SD14} \cite{nist14}, \emph{FVC2002 DB1\_A \& DB3\_A} \cite{fvc2002}, \emph{FVC2004 DB1\_A \& DB2\_A} \cite{fvc2004}, \emph{FVC2006 DB1\_A} \cite{fvc2006}) and an in-house dataset.
Experimental results show that the proposed algorithm outperforms previous state-of-the-art methods in both partial fingerprint verification and pose estimation, and also demonstrates high efficiency.

The main contributions of our paper can be summarized as:
\begin{itemize}
	\item We propose a novel approach for partial fingerprint recognition by jointly estimating the authentication probability and relative pose instead of the previous independent stages.
	\item A CNN-Transformer hybrid network structure is presented to aggregate their respective advantages in local feature extraction and global information interaction.
	\item A lightweight pre-training task on fingerprint enhancement is specifically designed to further improve the generalization ability of our proposed network.
	\item Extensive experiments are conducted on diverse datasets to evaluate representative state-of-the-art algorithms, which demonstrates the superiority of our proposed method.
\end{itemize}

The paper is organized as follows: 
Section \ref{sec:related_work} reviews the related works. 
Section \ref{sec:method} introduces the proposed partial fingerprint recognition algorithm.
Section \ref{sec:dataset} describes the details and usage of datasets.
Section \ref{sec:experiments} presents the experimental results and discussions.
Finally, we draw conclusions in Section \ref{sec:conclusion}.

\section{Related Work}\label{sec:related_work}
According to different feature representation forms, existing one-to-one fingerprint matching algorithms can be mainly divided into three categories:  key points based, fixed-length region descriptor based, and pairwise interrelated feature based.
In the first two types of algorithms highly generalized features are stored and compared here (in other words, each fingerprint undergoes feature extraction only once) to enable efficient searches in large-scaled galleries.
On the other hand, pairwise interrelated feature based methods could use richer information (features need to be re-extracted for each match) and is more suitable for confirmation scenarios with high precision requirements and few gallery\,/\,reference fingerprints, such as unlocking a mobile phone.
Given the multitude of fingerprint matching algorithms, this rough classification may not comprehensively encompass all works. 
Nonetheless, we still endeavor to summarize them to the best of our knowledge.
Fig. \ref{fig:intro_method} shows the corresponding schematic diagrams of these three types of matching algorithms and ours.

\subsection{Key Points Based}
Minutiae are the most popular features used for fingerprint matching \cite{maltoni2022handbook}.
Researchers define minutiae descriptors using the attributes of minutiae themselves \cite{cappelli2010minutia} and auxiliary information (such as orientation \cite{chen2009reconstructing}, period map \cite{feng2008combining}, ridge \cite{choi2011fingerprint} etc.).
Considering the lack of reliable minutiae in latent fingerprints, Cao \etal \cite{cao2019end} uniformly sample images into small patches and used a network to extract fixed length vector to describe virtual minutiae.
However, as explained in Fig. \ref{fig:intro_ex}, these features are not discriminative enough in partial fingerprints.
On the other hand, some studies introduce universal key points in computer vision field into fingerprints \cite{yamazaki2015sift,mathur2016methodology}.
Features extracted in this way are more densely distributed, but at the same time more sensitive to unreliable image details and therefore easily confused.
There are also algorithms that exploit Level-3 features from high-resolution images \cite{teixeira2017new,zhang2019combining}, which are relatively expensive to deploy in most commercial applications.
After getting key point sets, correspondence between them will be established based on the similarity of features and spatial structures, and then relative poses and matching scores are calculated.

\subsection{Fixed-length Region Representation Based}
Early studies use hand-crafted statistics to characterize fingerprint regions and convert them into fixed length vectors \cite{jain1999fingercode, nanni2008local,cappelli2011fast}. 
This feature form significantly improves the speed of large database indexing at the expense of reduced accuracy, and is usually used for rough searching.
Engelsma \etal \cite{engelsma2021learning} propose an end-to-end network to extract fixed-length global representations of minutiae and textures, which greatly improves the performance of fixed-length representation methods.
Inspired by this, researchers have conducted more exploration and improvement of fixed-length representation methods based on deep learning \cite{grosz2022c2cl,wu2022minutiae,duan2023fingerprint,grosz2024afr}.
In addition, Gu \etal \cite{gu2021latent} conduct dense sampling on complete fingerprints and compare the fixed length descriptors of each patch, ultimately making a comprehensive decision.
These algorithms usually require pose rectification in advance, such as adaptively adjusting images through a spatial transformation layer \cite{engelsma2021learning,grosz2024afr}, rectifying absolute pose of fingerprints through a separate network \cite{duan2023fingerprint}, or exploiting the positioning relationships implicit in minutiae \cite{wu2022minutiae}.
The error of pose estimation is tolerated to a certain extent because the extracted features are a generalization of regional characteristics.
However, current methods may struggle to address partial fingerprints, as the pose estimation on them is often unstable.

\subsection{Pairwise Interrelated Feature Based}
Compared with key points or fixed-length descriptors, gray-level features of pixels contain raw information about local textures.
Traditional algorithms use image correlation for fingerprint verification \cite{nandakumar2004local,lindoso2007increasing}, which can be affected by skin distortion and has limited discriminating ability.
Dense fingerprint rectification and registration can reduce the impact of skin distortion on image correlation \cite{guan2023regression,guan2024phase}, but simple correlation cannot separate genuine and impostor matches very well.
With the development of deep learning, researchers apply neural networks to directly extract and verify interrelated features from two input images, significantly improving the verification performance \cite{zeng2019research,bakhshi2019end,liu2020novel,he2022pfvnet}.
Specifically, He \etal \cite{he2022pfvnet} use multi-rotation and multi-size cropped image pairs as combined inputs to further assist the network in understanding scenarios with large spatial transformations or small overlapping areas.
It should be noted that a significant difference between this type of method and methods based on fixed-length region descriptor is that the relative pose of a certain fingerprint is rectified instead of the respective absolute pose.
Features of such methods need to be compared with precise spatial correspondence and thus have stricter requirements for alignment.
Obviously, the verification performance will significantly decline with inaccurate pose estimation \cite{he2022pfvnet}.

\begin{figure*}[!t]
	\begin{center}
		\includegraphics[width=.95\linewidth]{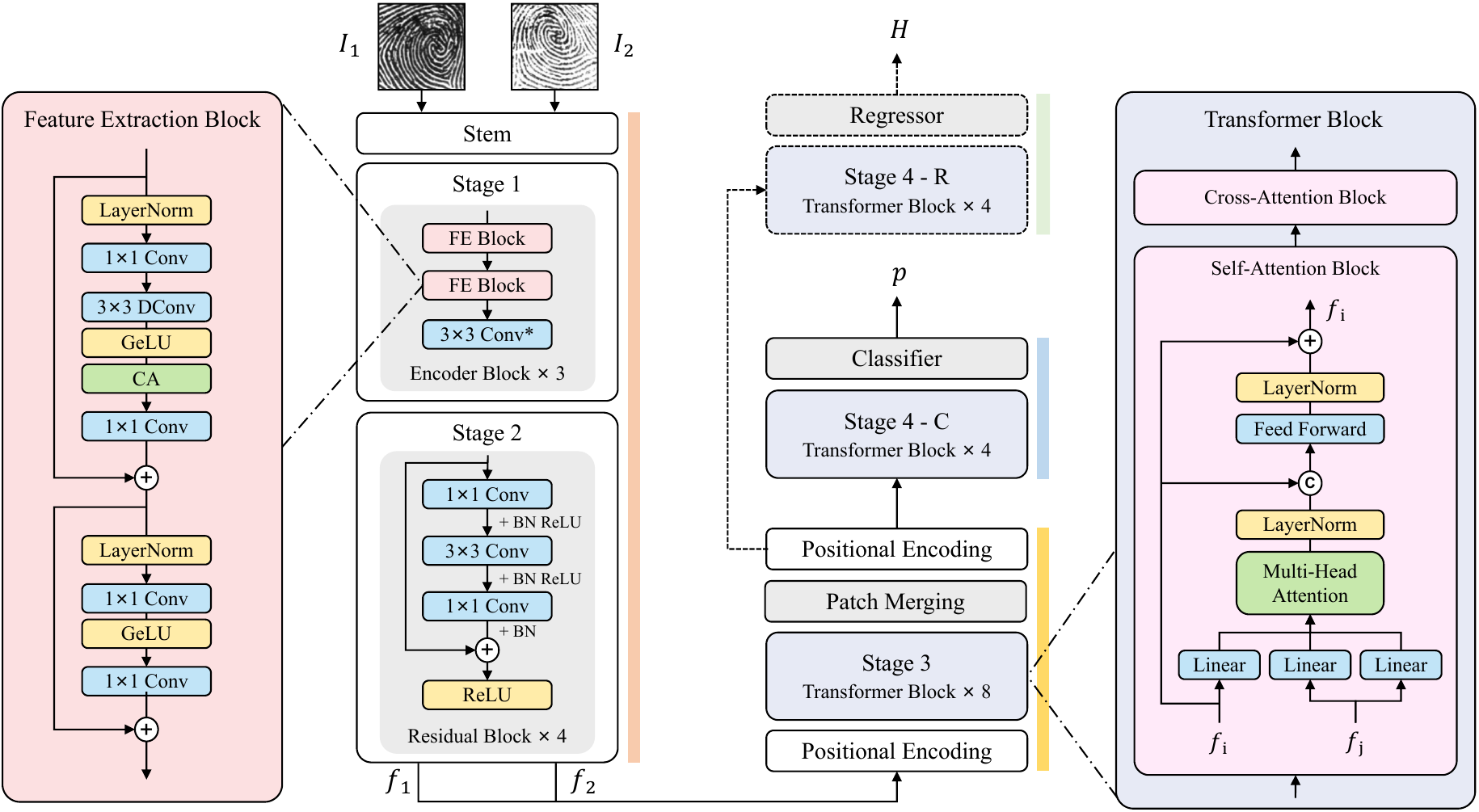}
	\end{center}
	\caption{An overview of JIPNet.
	{
	Paired fingerprint patches with the same shape are input, specifically $160 \times 160$, $120 \times 120$, or $96 \times 96$ in this paper.
	The outputs `\emph{p}' and `\emph{H}' of respective task heads correspond to the classification probability (whether the input fingerprints come from the same finger) and rigid transformation parameters (relative translation and rotation) respectively.
	Detailed structure is shown in Table \ref{tab:network}.
	} 
	Bars are presented on the left to indicate each phase, where the color definition refers to Fig. \ref{fig:intro_method}. 
	The process passed by parallel arrows indicates that the corresponding modules share weights and their functions can be executed in parallel. 
	* represents the number of channels are doubled after the corresponding convolution.
	Dotted parts are only used to assist training and could be pruned in practical verification tasks. 
	}
	\label{fig:network}
\end{figure*}

\begin{table}[!t]
	\belowrulesep=0pt
	\aboverulesep=0pt
	\renewcommand\arraystretch{1.3}
	\caption{{
	Detailed configurations of JIPNet. A pair of $160 \times 160$ fingerprint patches are inputted as an example.}}
	\label{tab:network}
	\vspace{-0.4cm}
	
	\begin{center}
		\begin{threeparttable}
			\begin{tabular}{c|c|c|c}
				\toprule
				Phase & Layer & Operator & Output shape \\
				\midrule
				\multirow{6}{*}{\makecell[c]{Independent\\Feature\\Extraction}} 
				& Stem & $3\times3$ Conv & $32, 160 \times 160$ \\
				\cline{2-4}
				{} & \multirow{3}{*}{\makecell[c]{Stage 1}} 
				& Encoder Block & $64, 80 \times 80$ \\
				\cline{3-4}
				{} & {} 
				& Encoder Block & $128, 40 \times 40$ \\
				\cline{3-4}
				{} & {} 
				& Encoder Block & $256, 20 \times 20$ \\
				\cline{2-4}
				{} & \multirow{2}{*}{\makecell[c]{Stage 2}} 
				& \multirow{2}{*}{\makecell[c]{Residual Block \\ $\times 4$}}
				& \multirow{2}{*}{\makecell[c]{$264, 20 \times 20$}} \\
				{} & {} & {} & {} \\
				\hline
				\multirow{5}{*}{\makecell[c]{Interrelated\\Feature\\Extraction}} 
				& \multicolumn{2}{c|}{Positional Encoding} & $264, 400$\\
				\cline{2-4}
				{} & \multirow{2}{*}{\makecell[c]{Stage 3}}
				& \multirow{2}{*}{\makecell[c]{Transformer Block \\ $\times 8$}} 
				& \multirow{2}{*}{\makecell[c]{$264, 400$}} \\
				{} & {} & {} & {} \\
				\cline{2-4}
				{} & \multicolumn{2}{c|}{Patch Merging} & $384, 100$\\
				\cline{2-4}
				{} & \multicolumn{2}{c|}{Positional Encoding} & $384, 100$\\
				\hline
				\multirow{4}{*}{\makecell[c]{Identity\\Verification}} 
				& \multirow{3}{*}{\makecell[c]{Stage 4 - C}} 
				& \multirow{2}{*}{\makecell[c]{Transformer Block \\ $\times 4$}}
				& \multirow{2}{*}{\makecell[c]{$384, 100$}}\\
				{} & {} & {} & {} \\
				\cline{3-4}
				{} & {} & Concatenate & $768, 100$\\
				\cline{2-4}
				{} & \multirow{2}{*}{\makecell[c]{Classifier}} & Linear Projection & $1, 1$\\
				\cline{3-4}
				{} & {} & Sigmoid & $1, 1$ \\
				\hline
				\multirow{4}{*}{\makecell[c]{Pose\\Prediction}} 
				& \multirow{3}{*}{\makecell[c]{Stage 4 - R}} 
				& \multirow{2}{*}{\makecell[c]{Transformer Block \\ $\times 4$}}
				& \multirow{2}{*}{\makecell[c]{$384, 100$}}\\
				{} & {} & {} & {} \\
				\cline{3-4}
				{} & {} & Concatenate & $768, 100$\\
				\cline{2-4}
				{} & Regressor & Linear Projection & $4, 1$\\
				\bottomrule
			\end{tabular}
		\end{threeparttable}
	\end{center}
\end{table}

\section{Method}\label{sec:method}
In this paper, we propose JIPNet,  a CNN-Transformer hybrid network, to jointly estimate the authentication probability and relative pose of partial fingerprint pairs.
Fig. \ref{fig:network} gives the complete flowchart of our proposed algorithm.
The overall process can be divided into four phases, which are indicated by different color bars in schematic: independent feature extraction (red), interrelated feature extraction (gold), identity verification (blue), and pose prediction (green).
For an input fingerprint pair, our algorithm first extracts the respective features in parallel using convolutional blocks with shared weights (in Stem and Stages 1, 2).
These two sets of features are then concatenated and fed to transformer blocks in Stage 3 (along with some extra microprocessing) for sufficient information interaction both locally and globally.
Finally, respective output layers utilize the aggregated information to predict classification probabilities for authentication (Stage 4\,-\,C and Classifier) and regression values for relative poses (Stage 4\,-\,R and Regressor).
It should be noted that these two tasks are performed in an integrated manner instead of independently as previous methods because we hope to utilize the mutually beneficial coupling relationship between each other, inspired by works on jigsaw puzzles \cite{paumard2020deepzzle,markaki2023jigsaw}.
In particular, pose-related regression layers could be pruned to improve efficiency since only identity identification is necessary in most applications.

\subsection{Independent Feature Extraction} \label{subsec:feature_extraction}

Previous studies have proven that early stacked convolution blocks can significantly improve the stability and performance of subsequent vision transformers \cite{xiao2021early,han2023survey}.
Similarly, we also use convolution blocks to parallelly extract independent features from each input partial fingerprint in the early stages of proposed network.

Specifically, the feature extraction module sequentially contains: (1) a $3 \times 3$ convolutional stem with stride $1$ (output $32$ channels); 
(2) stacked encoder blocks in Stage 2 for preliminary extraction of robust features, which consisting of two Feature Extraction (FE) blocks and a convolutional layer with stride $2$ for downsampling; 
(3) stacked residual blocks in Stage 3 to further enhance the expressive ability of neural network while mitigating gradient problems in deep layers.
Among them, FE block is introduced from a simple but efficient baseline for image restoration tasks \cite{chen2022simple}, which is formally similar to the combination of Mobile Convolution (MBConv) block \cite{sandler2018mobilenetv2} and Multilayer Perceptron (MLP) block \cite{dosovitskiy2020vit}, to improve the robustness of feature encoding in different image qualities.
The details of corresponding architecture is presented on the left side of Fig. \ref{fig:network}, where `DConv' and `CA' are depthwise convolution \cite{sandler2018mobilenetv2} and channel attention \cite{hu2018squeeze} respectively.

\begin{figure*}[!t]
	\centering
	\subfloat{\includegraphics[height=.37\linewidth]{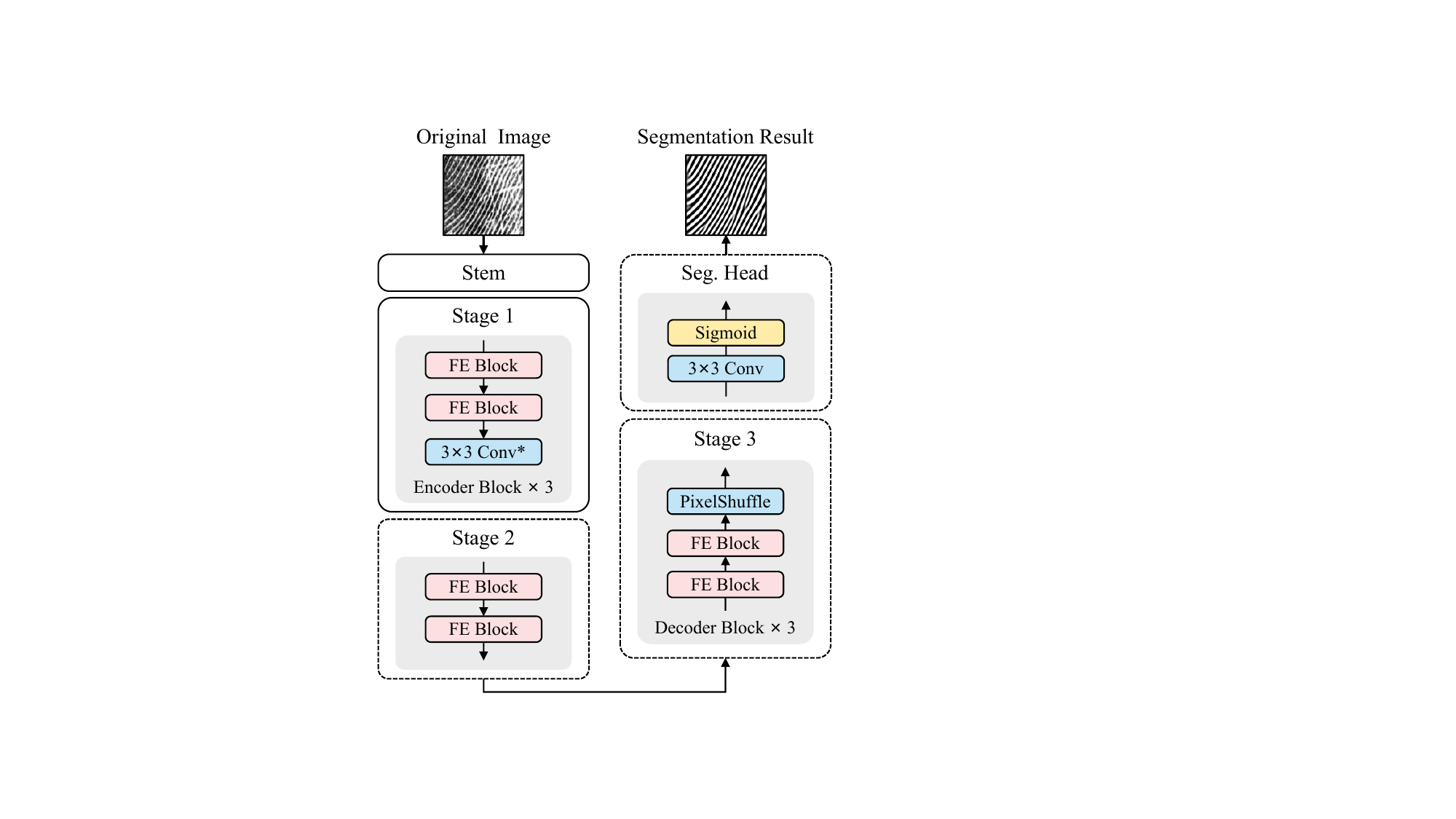}%
	}
	\hfil
	\subfloat{\includegraphics[height=.37\linewidth]{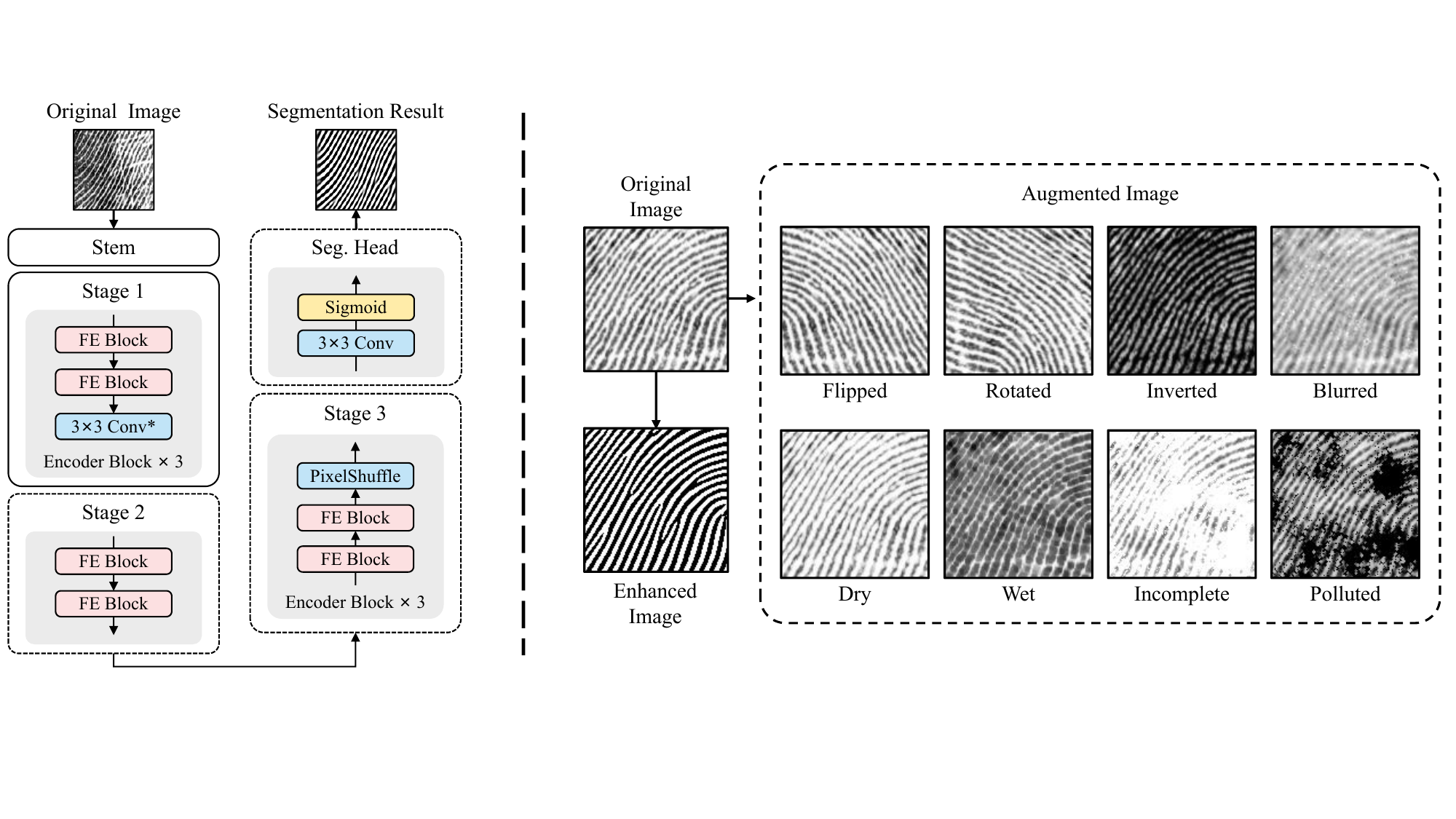}%
	}
	\hfil
	\subfloat{\includegraphics[height=.37\linewidth]{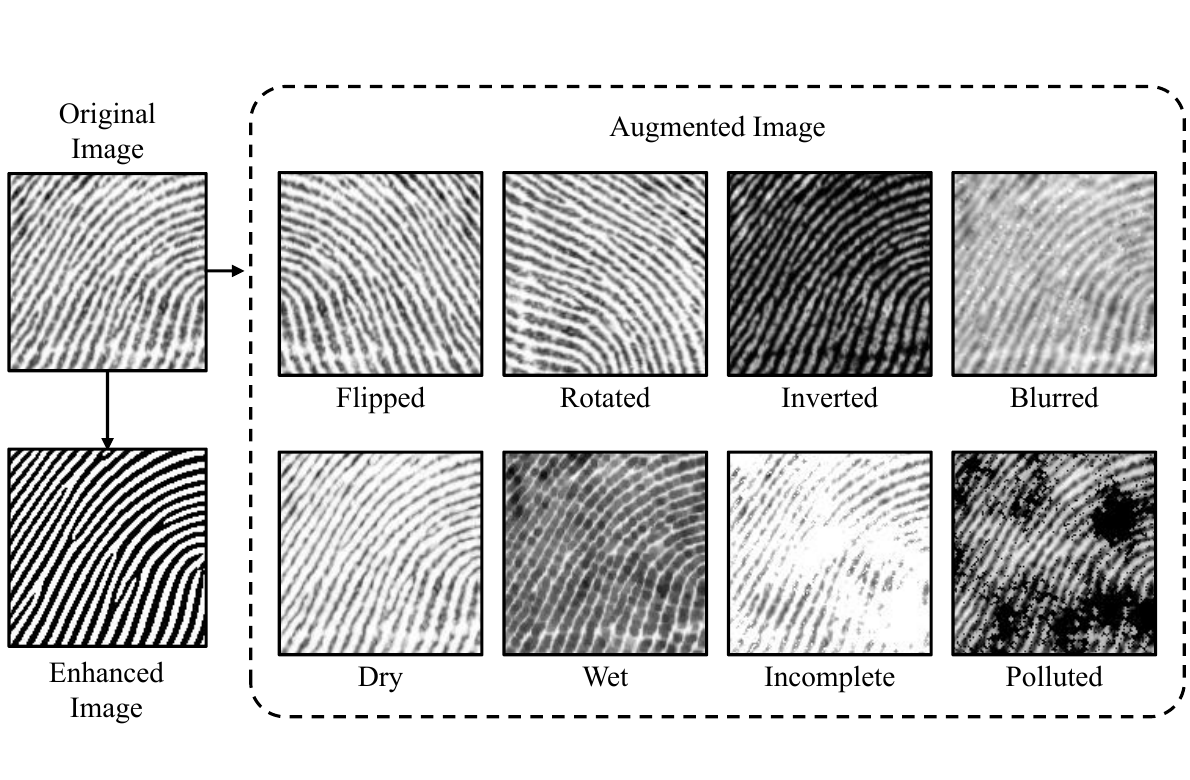}%
	}
	\vspace*{-1mm}
	\caption{Illustration of fingerprint enhancement task for pre-training. 
		The network architecture is shown on the left, where solid and dotted boxes indicate the pre-trained parameters will (or not) be loaded into corresponding modules in JIPNet.
	{
	Detailed structure is shown in Table \ref{tab:network_pretrain}.
	} 	
	The right subfigure gives representative examples of inputs (original \& augmented image) and targets (enhanced image) generation.}
	\label{fig:network_pretrain}
\end{figure*}

\begin{table}[!t]
	\belowrulesep=0pt
	\aboverulesep=0pt
	\renewcommand\arraystretch{1.3}
	\caption{{
			Detailed configurations of the fingerprint enhancement network in pre-training task. A pair of $160 \times 160$ fingerprint patches are inputted as an example.}}
	\label{tab:network_pretrain}
	\vspace{-0.4cm}
	
	\begin{center}
		\begin{threeparttable}
			\begin{tabular}{c|c|c}
				\toprule
				Layer & Operator & Output shape \\
				\midrule
				Stem & $3\times3$ Conv & $32, 160 \times 160$ \\
				\hline
				\multirow{3}{*}{\makecell[c]{Stage 1}} 
				& Encoder Block & $64, 80 \times 80$ \\
				\cline{2-3}
				{} 
				& Encoder Block & $128, 40 \times 40$ \\
				\cline{2-3}
				{} 
				& Encoder Block & $256, 20 \times 20$ \\
				\hline
				\multirow{2}{*}{\makecell[c]{Stage 2}} 
				& \multirow{2}{*}{\makecell[c]{FE Block \\ $\times 2$}}
				&  \multirow{2}{*}{\makecell[c]{$256, 20 \times 20$}} \\
				{} & {} & {} \\
				\hline
				\multirow{3}{*}{\makecell[c]{Stage 3}} 
				& Decoder Block & $128, 40 \times 40$ \\
				\cline{2-3}
				{} 
				& Decoder Block & $64, 80 \times 80$ \\
				\cline{2-3}
				{} 
				& Decoder Block & $32, 160 \times 160$ \\
				\hline
				\multirow{2}{*}{\makecell[c]{Seg. Head}} 
				& $3 \times 3$ Conv & $1, 160 \times 160$ \\
				\cline{2-3}
				{} & Sigmoid & $1, 160 \times 160$ \\
				\bottomrule
			\end{tabular}
		\end{threeparttable}
	\end{center}
\end{table}

\begin{figure}[!t]
	\centering
	\includegraphics[width=1\linewidth]{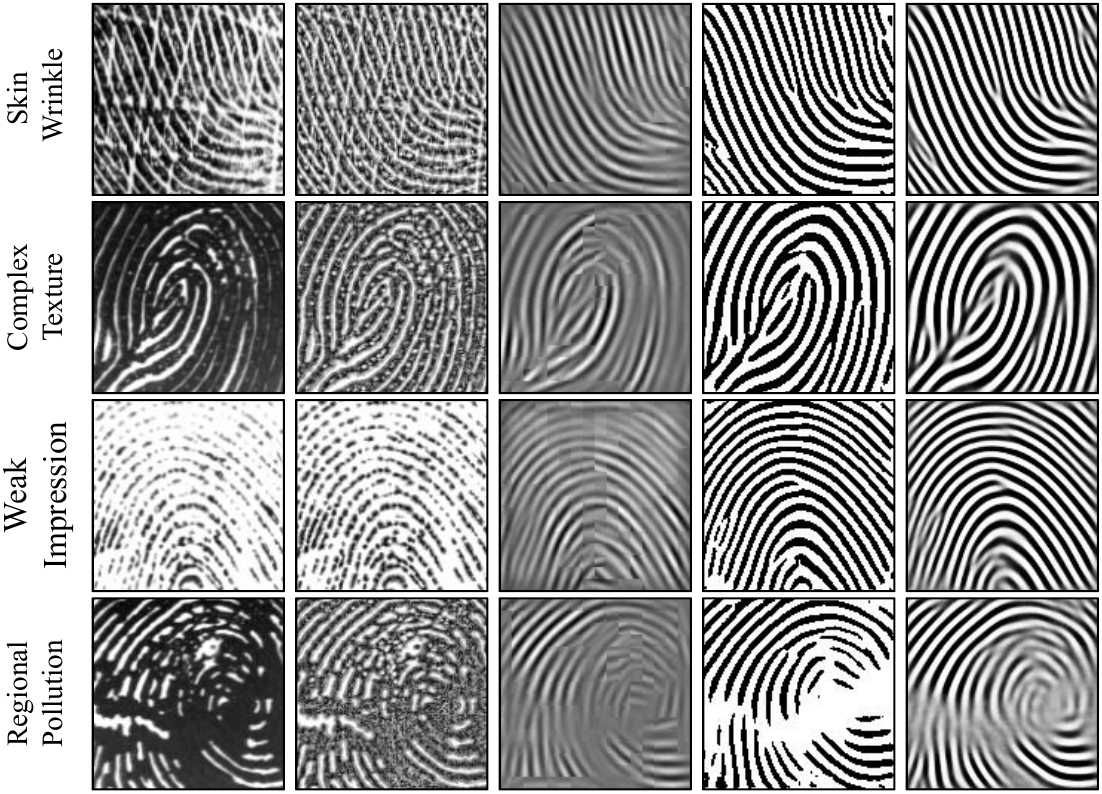}
	\vspace*{-5mm}
	\caption{Comparison of representation enhancement methods on four low-quality fingerprints.
		Each column from left to right is original image and corresponding enhancement results of CLAHE \cite{zuiderveld1994contrast}, FingerNet \cite{tang2017fingernet}, VeriFinger \cite{VeriFinger} and our proposed method. 
	}
	\label{fig:example_enh}
\end{figure}

Furthermore, we design a pre-training task and corresponding data augmentation strategies to help the network better extract and reconstruct the essential features of fingerprints.
As shown in Fig. \ref{fig:network_pretrain}, the pre-training network consists only of FE block and convolution without extra complex components, arranged in a U-shape for fingerprint enhancement follows \cite{grosz2023latent,zhu2023fingergan}.
For data preparation of inputs, we randomly crop small patches ($128 \times 128$ in this paper) from high-quality rolled fingerprints as prototypes to ensure sufficient texture patterns at each position while accelerating the training process.
Corresponding cases after augmentation are given on the right of Fig. \ref{fig:network_pretrain}, including:
\begin{itemize}
	\item Mirror flip, rotation (by 90, 180 or 270 degree) and grayscale inversion to increase data diversity;
	\item Gaussian blur and noise with different cores and variances to simulate irregular sensor noise;
	\item Dilation and erosion with structural elements of different sizes to simulate dry and wet fingers respectively;
	\item White blobs with random parameters which is added or subtracted to simulate fingerprint-specific noise in incomplete or contaminated scenarios \cite{cappelli2004improved}.
\end{itemize}
These augmentation strategies are randomly selected and combined to generate the final input image.
On the other hand, the corresponding binary fingerprint calculated by VeriFinger SDK 12.0 \cite{VeriFinger} is used as the enhancement target.
Parameters of Stem and Stage 1 are then loaded into corresponding modules of JIPNet.

Fig. \ref{fig:example_enh} shows representative examples of several typical enhancement methods.
It can be seen that our approach is smoother than methods based on grayscale mapping \cite{zuiderveld1994contrast} or filtering \cite{tang2017fingernet}.
In addition, our approach also shows certain robust performance when regional information is missing or wrong.
This qualitative comparison strongly proves that our pre-training strategy could help the network understand more essential and complete features, and also avoids additional cumbersome enhancement steps.
Further quantitative results are given in ablation experiments below.

\subsection{Interrelated Feature Extraction}
The hybrid strategy of using early CNN and subsequent transformer to aggregate local details and global information has achieved great success in multiple fields \cite{carion2020end, jiang2021cotr, shen2023semi}.
Inspired by these, we utilize transformers to capture long-range dependencies within and across paired features, thereby ensuring sufficient information interaction.
The paired features $f_1$ and $f_2$ are first biased with a sinusoidal positional encoding term \cite{carion2020end} as follows: 
\begin{equation}
	\begin{aligned}
		&P E_{(x,y, 4 i)}  =\sin \left(x / 10000^{4 i / d}\right) \;,\\
		&P E_{(x,y, 4 i + 1)}  =\cos \left(x / 10000^{4 i / d}\right) \;,\\
		&P E_{(x,y, 4 i + 2)}  =\sin \left(y / 10000^{4 i / d}\right)  \;,\\
		&P E_{(x,y, 4 i + 3)}  =\cos \left(y / 10000^{4 i / d}\right) \;,
	\end{aligned}
	\label{eq:positional_encoding}
\end{equation}
where $x$ and $y$ denote the 2D position, $i$ and $d$ denote the current and total dimensions.
Features with added absolute positional information are then spatially flattened into 1D sequences and fed into stacked transformers with interleaved self- and cross- attention.
It should be noted that the next two attention blocks have the same structure, including residual-connected Multi-Head Attention (MHA) and Feed-Forward Network (FFN) \cite{dosovitskiy2020vit}, and only the data streams are differentiated by their respective expected effects.
Specifically, let $\operatorname{SA}\left(f_i,f_j \right)$ (Self-Attention) and $\operatorname{CA}\left(f_i,f_j\right)$ (Cross-Attention) represent the corresponding sub-blocks in Fig. \ref{fig:network}, the complete process of transformer block is defined as:
\begin{equation}
	\begin{aligned}
		f_1^{\mathrm{S}}  = \operatorname{SA}\left(f_1,f_1 \right) & \;, \; f_2^{\mathrm{S}}  = \operatorname{SA}\left(f_2,f_2 \right) \;,\\
		f_1^{\mathrm{C}}  = \operatorname{CA}\left(f_1^{\mathrm{S}},f_2^{\mathrm{S}} \right) & \;,  \; f_2^{\mathrm{C}}  = \operatorname{CA}\left(f_2^{\mathrm{S}},f_1^{\mathrm{S}} \right) \;.\\
	\end{aligned}
	\label{eq:attention}
\end{equation}
The linear projection of $f_i$ and $f_j$ in MHA from left to right are $Query$, $Key$ and $Value$ matrices, which are then used for dot-product attention formally as:
\begin{equation}
	\operatorname{Attention}(Q, K, V)=\operatorname{softmax}\left(\frac{Q K^T}{\sqrt{d_k}}\right) V \;,
\end{equation}
where $d_k$ is the $Key$ dimention and the number of heads is fixed to $6$ at this stage.
In order to further condense features while reducing information loss, a patch merging layer \cite{liu2021swin} is subsequently incorporated for downsampling.
Finally, the positional encoding is applied again in the same way of Equation \ref{eq:positional_encoding}.

\subsection{Identification Verification \& Pose Prediction}
Similar to the previous phase, features are separately fed into stacked identical transformer blocks with head of $12$ to prepare for the corresponding task head.
We define identity verification as a binary classification task, that is, judging whether the input image pair comes from a same finger.
Let $\mathcal{F}$, $\operatorname{BN}$, $\operatorname{Swish}$ and $\operatorname{AvgPool}$ denote fully connected layer, batch normalization, swish activation and adaptive average pooling respectively, the classification probability $p$ is calculated as:
\begin{equation}
	\begin{aligned}
		\operatorname{MLP}[\cdot] &= \left\{\mathcal{F},\operatorname{BN},\operatorname{Swish} \;, \mathcal{F},\operatorname{BN},\operatorname{Swish}\right\} \;,\\
		\operatorname{Cla}[\cdot] &= \left\{\mathcal{F},\operatorname{AvgPool}, \operatorname{MLP}\right\} \;,\\
		p &= \operatorname{Sigmoid}(\operatorname{Cla}[f]) \;,
	\end{aligned}
	\label{eq:identity verification}
\end{equation}
where $f$ is the concatenated by the last $f_1^{\mathrm{C}}$ and $f_2^{\mathrm{C}}$ in channel dimension, operator $\{\,...\,\}$ means executing the contained functions sequentially from left to right.
On the other hand, we treat pose prediction as a regression task and predicting the rigid transformation parameters $H$ as:
\begin{equation}
	\begin{aligned}
		\operatorname{Reg}[\cdot] &= \left\{\mathcal{F},\operatorname{AvgPool}, \operatorname{MLP}\right\} \;,\\
		H &= \operatorname{Reg}[f] \;.
	\end{aligned}
	\label{eq:pose_prediction}
\end{equation}
Following previous work \cite{he2022pfvnet}, we represent the rotation relationship based on relative values of sine and cosine to circumvent the constraints on numerical range of angles, and directly describe the translation in terms of horizontal and vertical displacements.
With the prediction target of $H=[r_\mathrm{c},r_\mathrm{s},t_\mathrm{x},t_\mathrm{y}]$, the rigid alignment relationship of paired fingerprints can be modeled as:
\begin{equation}
	\begin{aligned}
		& \cos \theta={r_\mathrm{c}}/{\sqrt{r_\mathrm{c}^2+r_\mathrm{s}^2}} \;, \; \sin \theta={r_\mathrm{s}}/{\sqrt{r_\mathrm{c}^2+r_\mathrm{s}^2}} \;, \\[1mm]
		& 
		\left[\begin{array}{l}
			u_\mathrm{1} \\
			v_\mathrm{1} \\
			1
		\end{array}\right]=\left[\begin{array}{ccc}
		\cos \theta & -\sin \theta & t_\mathrm{x} \\
		\sin \theta & \cos \theta & t_\mathrm{y} \\
		0 & 0 & 1
		\end{array}\right] \cdot \left[\begin{array}{l}
			u_\mathrm{2} \\
			v_\mathrm{2} \\
			1
		\end{array}\right] \;,
	\end{aligned}
	\label{eq:rotation}
\end{equation}
where $\theta$ is the relative rotation angle, $(u, v)$ represents any corresponding position in images $1$ and $2$.

\subsection{Loss Function}
We optimize our network parameters through a comprehensive objective function consisting of two items. The total loss  can be formulated as follows:
\begin{equation}
	\mathcal{L} = \mathcal{L}_{\mathrm{cla}} + \lambda \cdot \mathcal{L}_{\mathrm{reg}} \;,
	\label{eq:loss}
\end{equation}
where $ \mathcal{L}_{\mathrm{cla}}$ and $\mathcal{L}_{\mathrm{reg}}$ are the identity classification term and pose regression term, respectively. 
The trade-off parameter $\lambda$ is set to $0.002$ in this paper.
Additionally, an extra image segmentation loss $ \mathcal{L}_{\mathrm{seg}}$ is proposed for fingerprint enhancement, which is applied only during pre-training.

\subsubsection{Identity Verification Loss}
For classification of identity verification, we adapt focal loss \cite{lin2017focal} to help our model better focus on difficult samples:
\begin{equation}
	\mathcal{L}_{\mathrm{cla}}=\left\{\begin{array}{c}
		-\alpha\left(1-p\right)^\gamma \log p, \text{ if } y=1 \\
		-\left(1-\alpha\right) p^\gamma \log (1-p), \text{ if } y=0
	\end{array}\right. \;,
	\label{eq:loss_cla}
\end{equation}
where $p$ represents the predicted probability, $y$ represents the 0-1 label of ground truth.
Balance factors $\alpha$ and $\gamma$ are set to $0.2$ and $2.0$ empirically.

\subsubsection{Pose Alignment Loss}
For regression of relate pose, weighted mean square error is utilzed for supervision:
\begin{equation}
	\begin{aligned}
		\mathcal{L}_{\mathrm{reg}}= \; & \omega \cdot \left[(\cos \theta^*-\cos \theta^{ })^2+(\sin \theta^*-\sin \theta^{ })^2\right] \\
		& +\left(1-\omega \right) \cdot \left[\left(t^*_\mathrm{x}-t^{ }_\mathrm{x} \right)^2+\left(t^*_\mathrm{y}-t^{ }_\mathrm{y}\right)^2 \right]
	\end{aligned}
	\label{eq:loss_reg}
\end{equation}
where the superscript * indicates the ground truth, otherwise the estimated result.
The relative angle $\theta$ and pixel level displacement $t$ is converted from the originally defined output form $H$ and Equation \ref{eq:rotation}.
Same as \cite{he2022pfvnet}, we fix $\omega$ to $0.99$ to balance the optimization of rotation and translation.

\subsubsection{Image Segmentation Loss}
We enhance the input fingerprints in form of binary segmentation, where the grayscale differences on modalities are minimized to make the texture pattern of ridges more prominent.
A simplified focal loss \cite{lin2017focal} is defined as the corresponding metric:
\begin{equation}
	\begin{aligned}
		\mathcal{L}_{\mathrm{seg}} & =-\frac{1}{|M|} \sum_M \left(1-q\right)^\gamma \log \left(q\right), \\
		q & =y \cdot p + \left(1 - y \right) \cdot \left(1-p\right) ,
	\end{aligned}
	\label{eq:loss_seg}
\end{equation}
where $p$ is the probability of foreground, $y$ is the ground truth of corresponding binary image obtained by VeriFinger \cite{VeriFinger}, and $M$ is the image mask.
The hyperparameter $\gamma$ is fixed to $2.0$ in experiments.

\section{Dataset Description}\label{sec:dataset}
\begin{figure*}[!t]
	\centering
	\subfloat[]{\includegraphics[height=.116\linewidth]{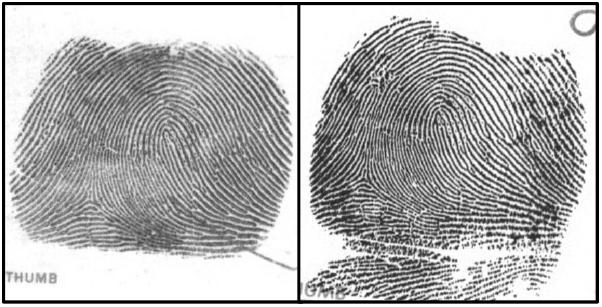}%
	\vspace{-1mm}}
	\hfil
	\subfloat[]{\includegraphics[height=.116\linewidth]{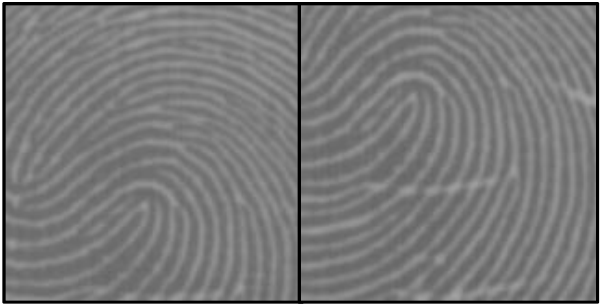}%
	\vspace{-1mm}}
	\hfil
	\subfloat[]{\includegraphics[height=.116\linewidth]{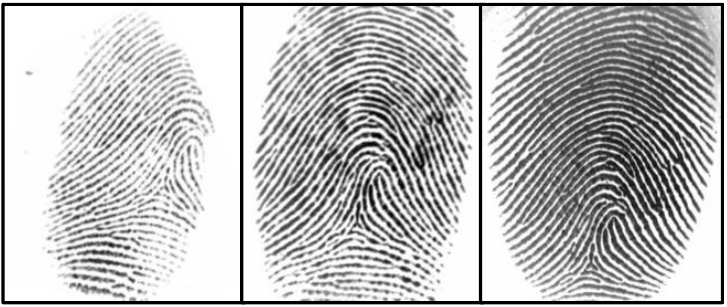}%
	\vspace{-1mm}}
	\hfil
	\subfloat[]{\includegraphics[height=.116\linewidth]{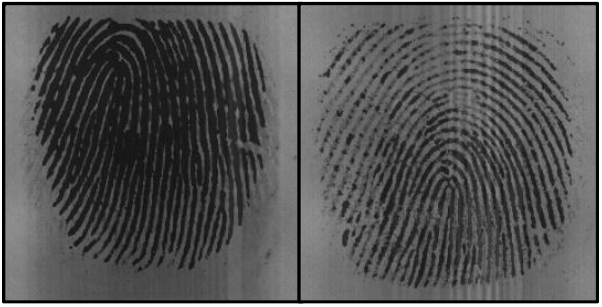}%
	\vspace{-1mm}}
	
	\par\vspace{2mm}
	
	\subfloat[]{\includegraphics[height=.116\linewidth]{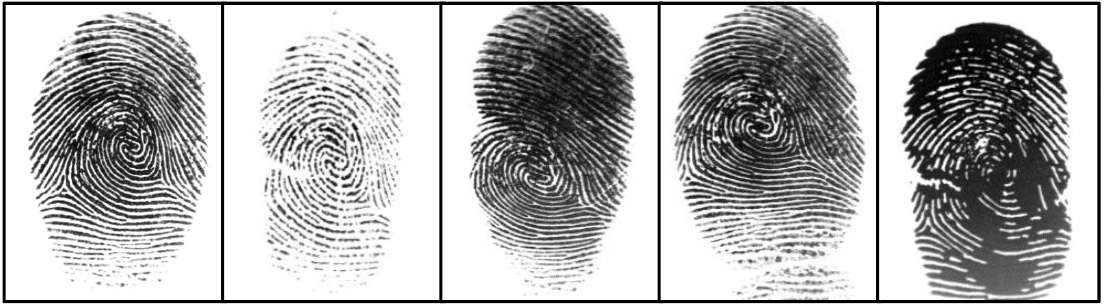}%
	\vspace{-1mm}}
	\hfil
	\subfloat[]{\includegraphics[height=.116\linewidth]{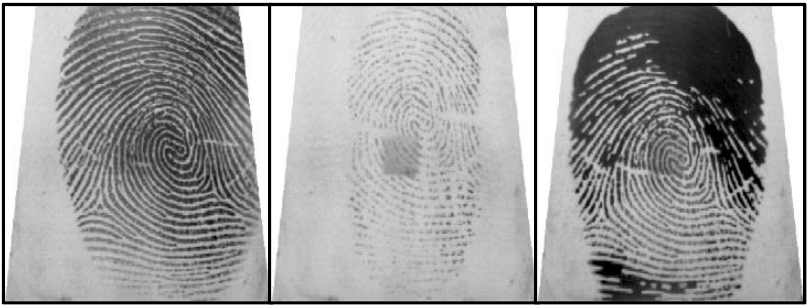}%
	\vspace{-1mm}}
	\hfil
	\subfloat[]{\includegraphics[height=.116\linewidth]{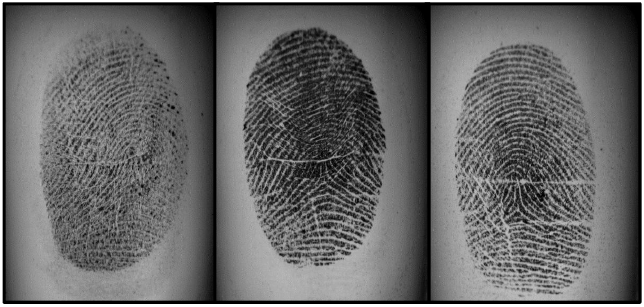}%
	\vspace{-1mm}}
	\caption{Image examples from different fingerprint datasets (a) NIST SD14 \cite{nist14}, (b) THU Small, (c) FVC2002 DB1\_A \cite{fvc2002}, (d) FVC2002 DB3\_A \cite{fvc2002}, (e) FVC2004 DB1\_A \cite{fvc2004}, (f) FVC2004 DB2\_A \cite{fvc2004}, (g)
		FVC2006 DB2\_A \cite{fvc2006}.}
	\label{fig:example_db}
\end{figure*}

Comprehensive evaluations are conducted on extensive public datasets,  including the representative public benchmarks \emph{NIST SD14} \cite{nist14},  \emph{FVC2002 DB1\_A \& DB3\_A} \cite{fvc2002}, \emph{FVC2004 DB1\_A \& DB2\_A} \cite{fvc2004}, and \emph{FVC2006 DB2\_A} \cite{fvc2006}.
In addition, we also use a private partial fingerprint dataset \emph{THU Small} to further examine the performance in capacitive smartphone situations.
Table \ref{tab:dataset} presents a detailed description about the composition and usage of these datasets.
In particular, we merge several fingerprint databases into a larger dataset, called the \emph{Hybrid Database (Hybrid DB)} , to make the implementation of various experiments more convenient and sufficient.
On the other hand, we further evaluate the performance of cross-domain generalization on other separate datasets.
Rolled fingerprint is utilized because it covers a more complete area at different pressing poses than plain fingerprints.
Image examples of different datasets are shown in Fig. \ref{fig:example_db}.
It can be seen that datasets selected in our experiments contains diverse fingerprint impressions with various modalities (normal, dry, wet, polluted, wrinkled, etc.).

\begin{table*}[!t]
	\renewcommand\arraystretch{1.3}
	\caption{All Fingerprint Datasets Used in Experiments.}
	\label{tab:dataset}
	\vspace{-0.4cm}
	\begin{center}
		\begin{threeparttable}
			\tabcolsep=9pt
			\begin{tabular}{ccccccc}
				\toprule
				\textbf{Type}           
				& \textbf{Dataset} 
				& \textbf{Scanners}              
				& \textbf{Description}        
				& \textbf{Usage} 
				& \textbf{Genuine pairs}
				& \textbf{Impostor pairs} \\
				\midrule
				Rolled
				& NIST SD14\tnote{\,a} \:\cite{nist14}
				& Inking
				& 27,000 fingers $\times$ 2 impressions
				& train \& test\tnote{\,c} 
				& 108,000
				& 108,000\\
				\midrule
				\multirow{2}{*}[-9mm]{Plain}
				& THU Small\tnote{\,b}
				& Capacitive
				& 100 fingers $\times$ 8 impressions
				& test
				& 5,600
				& 9,900\\
				& FVC2002 DB1\_A \cite{fvc2002}
				& Optical
				& 100 fingers $\times$ 8 impressions
				&  test
				& 5,600
				& 5,600\\
				& FVC2002 DB3\_A \cite{fvc2002}
				& Capacitive
				& 100 fingers $\times$ 8 impressions
				& test
				& 5,600
				& 5,600\\
				& FVC2004 DB1\_A \cite{fvc2004}
				& Optical
				& 100 fingers $\times$ 8 impressions
				& train \& test\tnote{\,c}
				& 22,400
				& 22,400\\
				& FVC2004 DB2\_A \cite{fvc2004}
				& Optical
				& 100 fingers $\times$ 8 impressions
				& train \& test\tnote{\,c} 
				& 22,400
				& 22,400\\
				& FVC2006 DB2\_A \cite{fvc2006}
				& Optical
				& 140 fingers $\times$ 12 impressions
				& train \& test\tnote{\,c}
				& 73,920
				& 73,920\\
				
				\bottomrule
			\end{tabular}
			\begin{tablenotes}
				\item[a] NIST SD14 was publicly available but later removed from public domain by NIST.
				\item[b] Corresponding dataset is private. Fingerprints are entered by volunteers in any comfortable pressing pose, with no special declaration or device restrictions. That is, there are quite a few genuine matching pairs with little or no overlap.
				\item[c] Datasets are merged and proportionally divided for training ($95\,\%$, called {Hybrid DB\_A}) and testing ($5\,\%$, called {Hybrid DB\_B}).  Identities between these two subsets are completely isolated.
			\end{tablenotes}
		\end{threeparttable}
	\end{center}
\end{table*}

\subsection{Parital Fingerprint Pair Simulation}
Considering the lack of publicly available datasets of partial fingerprints, image patches of specified sizes are synthesized  from rolled or plain datasets, similar to previous works \cite{he2022pfvnet,chen2022query2set}.
For any two fingerprints from the same finger, we first use VeriFinger \cite{VeriFinger} to align them based on matching minutiae.
Alignment in this way is accurate in most cases because original image pairs generally have large enough overlap areas of high quality.
Subsequently, we calculated the common mask of aligned fingerprints and randomly select the patch center of one partial fingerprint from it.
Taking this center as the zero point, another patch center is uniformly sampled in polar coordinates with a random angle ranging from $-180^\circ$  to $180^\circ$ and random radius ranging from 0 pixel to 100\,/\,70\,/\,20 pixels.
Finally, full fingerprints are cropped into partial images of $160\times160$, $128\times128$ and $96\times96$ with a  random relative rotation from $[-180^\circ,180^\circ]$, aiming to comprehensively evaluate the performance under different partial fingerprint sizes.
The parameters of rigid translation and rotation are converted to ground truth $H$ according to Equation \ref{eq:rotation}.
It should be noted that each pair of partial fingerprints are cropped from different original images to avoid possible information leakage caused by the same grayscale relationship or distortion pattern.

\subsection{Matching Protocols}
In order to balance the number of genuine\,/\,impostor matches and different fingerprint types\,/\,modalities, we implemented specific matching protocols during training and testing.
For the \emph{FVC} series, each fingerprint is combined with all mated fingerprints (with order) as genuine match pairs, and the first image of each pair is combined with a random impression of another finger as a new impostor match pair.
Taking \emph{FVC2002 DB1\_A} as an example, a total of $8\times7\times100\times2=11,200$ image pairs are combined in this way.
This protocol is also applied to \emph{NIST SD14} with a slightly different that symmetric matches are avoided, considering the large number of rolled images.
All combined pairs of full fingerprints are then randomly cropped to simulate several partial fingerprint scenes, with the number of executions set to $4$ (NIST SD14, FVC2004 DB1\_A \& DB2\_A, FVC2006 DB2\_A) for training \,/\, testing and $1$ (\emph{FVC2002 DB1\_A \& DB3\_A}) for testing.
As mentioned above, some representative datasets are merged as $\emph{Hybrid DB}$ and subsequently divided, secifically $430,768$ for training and $22,672$ for testing, which are distinguished by suffixes `\_A' and `\_B' in the following experiments.
The selection of genuine matches in real partial fingerprint dataset \emph{THU Small} also follows this principle, while pairing the first impression of every finger with each other as impostor pairs.
All detailed usage information is listed in corresponding footnotes of Table \ref{tab:dataset}.

\section{Experiments}\label{sec:experiments}
In this section, several representative state-of-the-art fingerprint algorithms are compared with our proposed approach, including:
\begin{itemize}
	\item \textbf{A-KAZE} \cite{mathur2016methodology}, a simple but comparative partial fingerprint recognition algorithm based on commonly used key points;
	\item \textbf{VeriFinger} SDK 12.0 \cite{VeriFinger}, a widely used,  top performing commercial software mainly based on minutiae;
	\item \textbf{DeepPrint} \cite{engelsma2021learning}, a global fixed-length representation extracted by deep network, which is highly influential;
	\item \textbf{DesNet} \cite{gu2021latent}, a descriptor network mainly designed for latent fingerprint, used to extract localized deep descriptors of densely sampled patches;
	\item {
		\textbf{AFR-Net} \cite{grosz2024afr}, an attention-driven fingerprint recognition network that extracts complementary global representations through ViT and ResNet embeddings.}
	\item \textbf{PFVNet} \cite{he2022pfvnet}, a state-of-the-art partial fingerprint verification algorithm recently proposed, which introduces multi-level features fusion and local self-attention mechanism.
\end{itemize}
{
In particular, since the effective information of partial fingerprints is significantly reduced compared to full-size fingerprints, the original pose rectification pre-step adopted in DeepPrint \cite{engelsma2021learning}, DesNet \cite{gu2021latent} and AFR-Net \cite{grosz2024afr} cannot be performed as expected.
Therefore, in this paper we use AlignNet, part of PFVNet \cite{he2022pfvnet}, to estimate and correct the relative pose of image pairs as substitute, which is marked by subscript `*'.}

We conducted extensive experiments to thoroughly evaluate the performance of above algorithms under partial fingerprints at different sizes, which in terms of score distribution, matching performance, alignment accuracy, visual analysis and efficiency.
In addition, ablation studies are demonstrated to validate the effectiveness of corresponding modules and strategies proposed in this paper.

\subsection{Implementation Details}
All partial fingerprint training processes are performed on the generated dataset \emph{Hybrid DB\_A}  with an initial learning rate of $1\mathrm{e}{-3}$ (end of $1\mathrm{e}{-6}$), cosine annealing scheduler, default AdamW optimizer and batch size of $128$ until convergence (about $12$ epochs).
The pre-training task proposed in Section \ref{subsec:feature_extraction} is trained for $200$ epochs using a subset of $5,000$ rolled images with corresponding random augmentation.
On the other hand, we completely reimplemented fixed-length representation based methods \cite{engelsma2021learning,gu2021latent} on full-size fingerprints in corresponding manner.
\emph{Hybrid DB\_B}, which has the same source as training set, is used to evaluate the performance under same data distribution.
Besides, scenarios of other image sizes and datasets are directly tested without fine-tuning to reflect the cross-domain adaptability.

\subsection{Score Distribution} \label{subsec:score_distribution}

\begin{figure}[!t]
	\centering
	\includegraphics[width=1\linewidth]{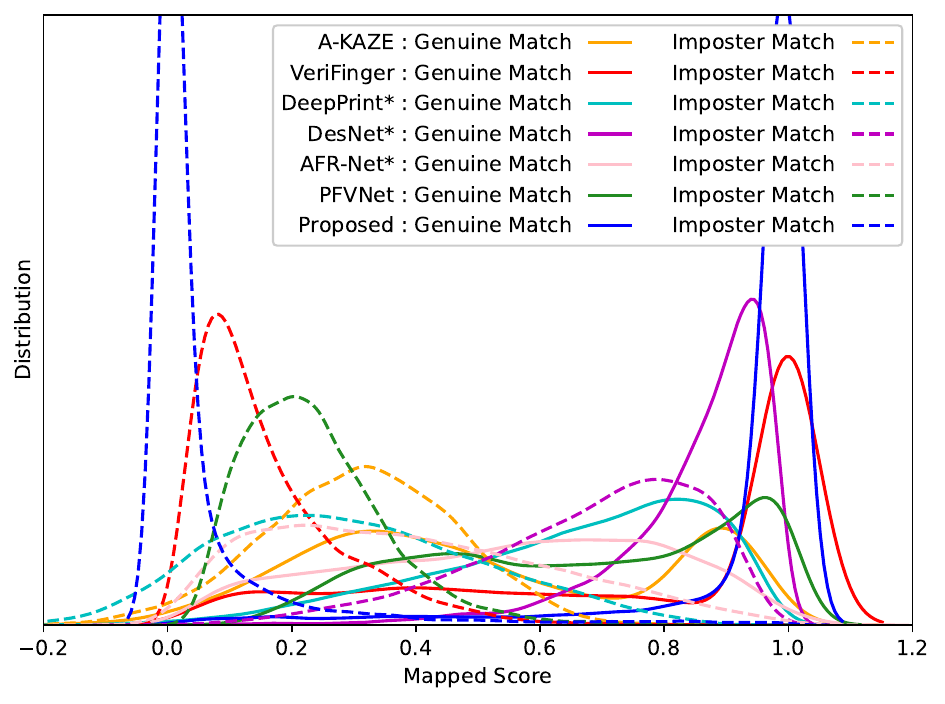}
	\vspace*{-8mm}
	\caption{
		{
			Probability density distributions of genuine and impostor matching scores on {Hybrid DB\_B} with image size $160\times160$. Scales on vertical axis are hidden because we are more concerned with the relative values.
		}
	}
	\label{fig:ex_score_distribution}
\end{figure}

We first present the score distribution of genuine and impostor matches on \emph{Hybrid DB\_B} with image size $160\times160$ to qualitatively demonstrate the discriminative capabilities of different algorithms.
Corresponding curves are shown in Fig. \ref{fig:ex_score_distribution}, where the scores of A-KAZE \cite{mathur2016methodology}, VeriFinger \cite{VeriFinger}, and DesNet \cite{gu2021latent} are linearly mapped to appropriate ranges that approximate $[0,1]$ for intuitive comparison.
A more separated pairwise distribution (i.e., fewer overlapping areas) indicates that the corresponding algorithm can better distinguish whether a certain fingerprint pair comes from the same finger.

\begin{table}[!t]
	\renewcommand\arraystretch{1.3}
	\caption{{
			Probability Distribution Distance Between Genuine and Impostor Scores on {Hybrid DB\_B} with Image Size $160\times160$.
	}}
	\label{tab:ex_score_distribution}
	\vspace{-0.4cm}
	
	\begin{center}
		\begin{threeparttable}
			\tabcolsep=15pt
			\begin{tabular}{lccc}
				\toprule
				\textbf{Method} \qquad & \textbf{JS} & \textbf{EMD} & \textbf{MMD} \\
				\midrule
				A-KAZE \cite{mathur2016methodology} \qquad & 0.42 & 0.20 & 0.72 \\
				VeriFinger \cite{VeriFinger} \qquad & 0.70 & 0.59 & 3.21 \\
				DeepPrint* \cite{engelsma2021learning} \quad & 0.59 &  0.36 & 2.03  \\
				DesNet* \cite{gu2021latent} \qquad &  0.47 & 0.17 & 1.10 \\
				AFR-Net* \cite{grosz2024afr} \qquad &  0.26 & 0.15 & 0.43 \\
				PFVNet \cite{he2022pfvnet} \qquad & 0.73 & 0.43 & 3.01 \\
				\rowcolor{black!10}
				Proposed \qquad & \textbf{0.89} & \textbf{0.87} & \textbf{6.36} \\
				\bottomrule
			\end{tabular}
		\end{threeparttable}
	\end{center}
\end{table}

Moreover, three common indicators are further computed and reported in Table \ref{tab:ex_score_distribution} to  quantitatively measure the differences between genuine and impostor probability distributions.
Among them, the Jensen–Shannon (JS) divergence is defined as: 
\begin{equation}
	\begin{aligned}
		\operatorname{KL}\left(p \| q\right) & =\int p(x) \log \frac{p(x)}{q(x)} d x \;,\\
		\operatorname{JS}\left(p \| q\right) & =\frac{1}{2} \operatorname{KL}\left(p \| \frac{p+q}{2}\right)+\frac{1}{2} \operatorname{KL}\left(q \| \frac{p+q}{2}\right) \;, \\
	\end{aligned}
	\label{eq:js}
\end{equation}
where $p$ and $q$ correspond to any two distributions.
Let $\Pi(p, q)$ represent all possible joint distributions of $p$ and $q$, $F$ represent the functions of sample space, the Earth-Mover's Distance (EMD) and Maximum Mean Discrepancy (MMD) are calculated by
\begin{equation}
	\operatorname{EMD}\left(p, q\right) =\inf _{\gamma \sim \Pi(p, q)} \mathbb{E}_{(x, y) \sim \gamma}[\|x-y\|] \;, 
	\label{eq:EMD}
\end{equation}
and
\begin{equation}
	\operatorname{MMD}\left(F, p, q\right)=\max _{f \in F}\left(E_{x \sim p}[f(x)]-E_{y \sim q}[f(y)]\right) \;. 
	\label{eq:MMD}
\end{equation}
Similarly, the larger these values, the farther between corresponding distribution pairs, which normally means a better identity verification performance.

According to the results in Fig. \ref{fig:ex_score_distribution} and Table \ref{tab:ex_score_distribution}, our method outperforms all other methods, while VeriFinger \cite{VeriFinger} and PFVNet \cite{he2022pfvnet} each have sub-optimal performance in some parts.
In particular, the distinguishing ability of DesNet* \cite{gu2021latent} is not ideal because it is designed based on the prior assumption of high-precision alignment, which is difficult to satisfy in current partial fingerprint situations.
{
Another noteworthy phenomenon is that the discriminative ability of AFR-Net \cite{grosz2024afr} is not satisfactory in this small image patch scenario,  despite its advanced performance on rolled/plain fingerprint matching.
We believe that its attention driven architecture effectively increases the ability to express local textures. 
However, this high sensitivity may actually limit robustness when there is limited available information (small effective fingerprint area) or excessive interference (limited overlap ratio).}

\subsection{Matching Performance}

\begin{table*}[!ht]
	\renewcommand\arraystretch{1.5}
	\belowrulesep=-0.2pt
	\aboverulesep=-0.2pt
	\caption{
	Matching Performance of State-of-the-Art Algorithms and Our Approach on Four Representative Datasets of Partial Fingerprint with Three Different Sizes.}
	\label{tab:ex_matching_performance}
	\vspace{-0.4cm}
	\begin{center}
		\begin{threeparttable}
				\tabcolsep=4.5pt
				\begin{tabular}{c | c  c | cccc | cccc | cccc | cccc}
					\toprule
					\multirow{2}{*}[-.5mm]{\textbf{Size}} 
					& \multirow{2}{*}[-.5mm]{\textbf{Sketch}} 
					& \multirow{2}{*}[-.5mm]{{\textbf{Method}}} 
					& \multicolumn{4}{c|}{{\textbf{Hybrid DB\_B}}} 
					& \multicolumn{4}{c|}{{\textbf{FVC2002 DB1\_A}}}
					& \multicolumn{4}{c|}{{\textbf{FVC2002 DB3\_A}}}
					& \multicolumn{4}{c}{{\textbf{THU Small}}} 
					\\
					\cmidrule(lr){4-7}\cmidrule(lr){8-11}\cmidrule(lr){12-15}\cmidrule(lr){16-19}
					{} & {} & {}
					&  \scriptsize\textbf{ACC}&  \scriptsize\textbf{AUC} &  \scriptsize\textbf{TAR} &  \scriptsize\textbf{EER}
					&  \scriptsize\textbf{ACC}&  \scriptsize\textbf{AUC} &  \scriptsize\textbf{TAR} &  \scriptsize\textbf{EER}
					&  \scriptsize\textbf{ACC}&  \scriptsize\textbf{AUC} &  \scriptsize\textbf{TAR} &  \scriptsize\textbf{EER}
					&  \scriptsize\textbf{ACC}&  \scriptsize\textbf{AUC} &  \scriptsize\textbf{TAR} &  \scriptsize\textbf{EER}
					\\
					\midrule
					\multirow{7}{*}{\makecell[c]{160\\$\times$\\160}
					}
					& {Fig. \ref{fig:intro_method_a}} & A-KAZE \cite{mathur2016methodology} 
					& 0.67 & 0.70 & 0.28 & 0.37 
					& 0.82 & 0.86 & 0.56 & 0.22 
					& 0.73 & 0.77 & 0.36 & 0.31 
					& 0.89 & 0.87 & 0.68 & 0.21 \\
					{} & {}  & VeriFinger \cite{VeriFinger} 
					& 0.85 & 0.88 & 0.56 & 0.18 
					& 0.95 & 0.97 & 0.85 & 0.07 
					& 0.90 & 0.93 & \textbf{0.76} & 0.12 
					& 0.92 & 0.91 & \textbf{0.76} & 0.15   \\
					{} & Fig. \ref{fig:intro_method_b}  & DeepPrint* \cite{engelsma2021learning} 
					& 0.79 & 0.87 & 0.16 & 0.21 
					& 0.84 & 0.92 & 0.32 & 0.16 
					& 0.81 & 0.89 & 0.16 & 0.19 
					& 0.82 & 0.86 & 0.26 & 0.22  \\
					{} & {} & DesNet* \cite{gu2021latent} 
					& 0.74 & 0.81 & 0.05 & 0.27 
					& 0.79 & 0.87 & 0.16 & 0.21 
					& 0.76 & 0.84 & 0.08 & 0.25 
					& 0.77 & 0.78 & 0.20 & 0.29   \\
					{} & {} & AFRNet* \cite{grosz2024afr}
					& 0.64 & 0.67 & 0.01 & 0.37 
					& 0.71 & 0.77 & 0.03 & 0.29 
					& 0.65 & 0.70 & 0.01 & 0.35 
					& 0.65 & 0.62 & 0.01 & 0.40 \\
					{} & Fig. \ref{fig:intro_method_c}  & PFVNet \cite{he2022pfvnet} 
					& 0.86 & 0.94 & 0.47 & 0.14 
					& 0.90 & 0.96 & 0.61 & 0.10 
					& 0.90 & 0.96 & 0.39 & 0.11 
					& 0.84 & 0.87 & 0.47 & 0.22 \\
					{} & \cellcolor{black!10}{Fig. \ref{fig:intro_method_d}}  & \cellcolor{black!10}{Proposed\;} 
					& \cellcolor{black!10}\textbf{0.96} & \cellcolor{black!10}\textbf{0.99} & \cellcolor{black!10}\textbf{0.71} & \cellcolor{black!10}\textbf{0.04} 
					& \cellcolor{black!10}\textbf{0.99} & \cellcolor{black!10}\textbf{0.99} & \cellcolor{black!10}\textbf{0.93} & \cellcolor{black!10}\textbf{0.01} 
					& \cellcolor{black!10}\textbf{0.97} & \cellcolor{black!10}\textbf{0.99} & \cellcolor{black!10}{0.67} & \cellcolor{black!10}\textbf{0.03} 
					& \cellcolor{black!10}\textbf{0.92} & \cellcolor{black!10}\textbf{0.97} & \cellcolor{black!10}{0.70} & \cellcolor{black!10}\textbf{0.10}\\
					\midrule
					\multirow{7}{*}{\makecell[c]{128\\$\times$\\128}
					}
					& {Fig. \ref{fig:intro_method_a}} & A-KAZE \cite{mathur2016methodology} 
					& 0.60 & 0.63 & 0.22 & 0.39 
					& 0.69 & 0.73 & 0.45 & 0.27 
					& 0.63 & 0.67 & 0.20 & 0.34 
					& 0.79 & 0.74 & 0.40 & 0.34  \\
					{} & {}  & VeriFinger \cite{VeriFinger} 
					& 0.72 & 0.74 & 0.39 & 0.27 
					& 0.82 & 0.85 & 0.71 & 0.13 
					& 0.77 & 0.79 & 0.55 & 0.20 
					& 0.83 & 0.80 & 0.51 & 0.28  \\
					{} & Fig. \ref{fig:intro_method_b}  & DeepPrint* \cite{engelsma2021learning} 
					& 0.67 & 0.73 & 0.05 & 0.29 
					& 0.69 & 0.75 & 0.14 & 0.26 
					& 0.69 & 0.75 & 0.09 & 0.27 
					& 0.70 & 0.69 & 0.05 & 0.37  \\
					{} & {} & DesNet* \cite{gu2021latent} 
					& 0.64 & 0.69 & 0.03 & 0.32 
					& 0.66 & 0.71 & 0.05 & 0.29 
					& 0.64 & 0.69 & 0.04 & 0.32 
					& 0.70 & 0.67 & 0.06 & 0.38   \\
					{} & {} & AFRNet* \cite{grosz2024afr}
					& 0.54 & 0.54 & 0.01 & 0.45 
					& 0.56 & 0.57 & 0.01 & 0.40 
					& 0.54 & 0.54 & 0.01 & 0.43 
					& 0.63 & 0.52 & 0.01 & 0.49  \\
					{} & Fig. \ref{fig:intro_method_c}  & PFVNet \cite{he2022pfvnet}  
					& 0.73 & 0.81 & 0.16 & 0.24 
					& 0.71 & 0.79 & 0.18 & 0.25 
					& 0.73 & 0.82 & 0.16 & 0.24 
					& 0.73 & 0.74 & 0.14 & 0.33  \\
					{} & \cellcolor{black!10}{Fig. \ref{fig:intro_method_d}}  & \cellcolor{black!10}{Proposed\;} 
					& \cellcolor{black!10}\textbf{0.91} & \cellcolor{black!10}\textbf{0.97} & \cellcolor{black!10}\textbf{0.63} & \cellcolor{black!10}\textbf{0.05} 
					& \cellcolor{black!10}\textbf{0.94} & \cellcolor{black!10}\textbf{0.99} & \cellcolor{black!10}\textbf{0.89} & \cellcolor{black!10}\textbf{0.02} 
					& \cellcolor{black!10}\textbf{0.91} & \cellcolor{black!10}\textbf{0.97} & \cellcolor{black!10}\textbf{0.60} & \cellcolor{black!10}\textbf{0.05} 
					& \cellcolor{black!10}\textbf{0.88} & \cellcolor{black!10}\textbf{0.93} & \cellcolor{black!10}\textbf{0.53} & \cellcolor{black!10}\textbf{0.15}  \\
					\midrule
					\multirow{7}{*}{\makecell[c]{96\\$\times$\\96}
					}
					& {Fig. \ref{fig:intro_method_a}} & A-KAZE \cite{mathur2016methodology} 
					& 0.52 & 0.52 & 0.07 & 0.49 
					& 0.54 & 0.54 & 0.26 & 0.45 
					& 0.52 & 0.52 & 0.07 & 0.48 
					& 0.66 & 0.53 & 0.06 & 0.47  \\
					{} & {}  & VeriFinger \cite{VeriFinger} 
					& 0.59 & 0.60 & 0.25 & 0.31 
					& 0.65 & 0.66 & 0.53 & 0.20 
					& 0.62 & 0.63 & 0.26 & 0.24 
					& 0.74 & 0.67 & 0.25 & 0.37  \\
					{} & Fig. \ref{fig:intro_method_b}  & DeepPrint* \cite{engelsma2021learning} 
					& 0.61 & 0.64 & 0.05 & 0.30 
					& 0.59 & 0.61 & 0.01 & 0.30 
					& 0.61 & 0.64 & 0.08 & 0.27 
					& 0.64 & 0.56 & 0.02 & 0.45   \\
					{} & {} & DesNet* \cite{gu2021latent} 
					& 0.58 & 0.60 & 0.02 & 0.35 
					& 0.57 & 0.60 & 0.02 & 0.32 
					& 0.56 & 0.58 & 0.06 & 0.32 
					& 0.65 & 0.58 & 0.02 & 0.45    \\
					{} & {} & AFRNet* \cite{grosz2024afr}
					& 0.51 & 0.51 & 0.01 & 0.48 
					& 0.51 & 0.50 & 0.01 & 0.48 
					& 0.50 & 0.50 & 0.01 & 0.48 
					& 0.44 & 0.51 & 0.01 & 0.49   \\
					{} & Fig. \ref{fig:intro_method_c}  & PFVNet \cite{he2022pfvnet}  
					& 0.64 & 0.71 & 0.09 & 0.30 
					& 0.60 & 0.65 & 0.10 & 0.31 
					& 0.62 & 0.69 & 0.14 & 0.28 
					& 0.66 & 0.64 & 0.02 & 0.41 \\
					{} & \cellcolor{black!10}{Fig. \ref{fig:intro_method_d}}  & \cellcolor{black!10}{Proposed\;} 
					& \cellcolor{black!10}\textbf{0.79} & \cellcolor{black!10}\textbf{0.87} & \cellcolor{black!10}\textbf{0.48} & \cellcolor{black!10}\textbf{0.08} 
					& \cellcolor{black!10}\textbf{0.81} & \cellcolor{black!10}\textbf{0.89} & \cellcolor{black!10}\textbf{0.77} & \cellcolor{black!10}\textbf{0.04} 
					& \cellcolor{black!10}\textbf{0.79} & \cellcolor{black!10}\textbf{0.87} & \cellcolor{black!10}\textbf{0.63} & \cellcolor{black!10}\textbf{0.07} 
					& \cellcolor{black!10}\textbf{0.79} & \cellcolor{black!10}\textbf{0.82} & \cellcolor{black!10}\textbf{0.28} & \cellcolor{black!10}\textbf{0.26} \\
					\bottomrule
					
				\end{tabular}
			\begin{tablenotes}
				\item[] TAR represents TAR @ FAR = 1$\mathrm{e}$-3.
			\end{tablenotes}
		\end{threeparttable}
	\end{center}
\end{table*}

\begin{figure*}[!t]
	\centering
	\subfloat[A-KAZE \cite{mathur2016methodology}]{\includegraphics[width=.23\linewidth]{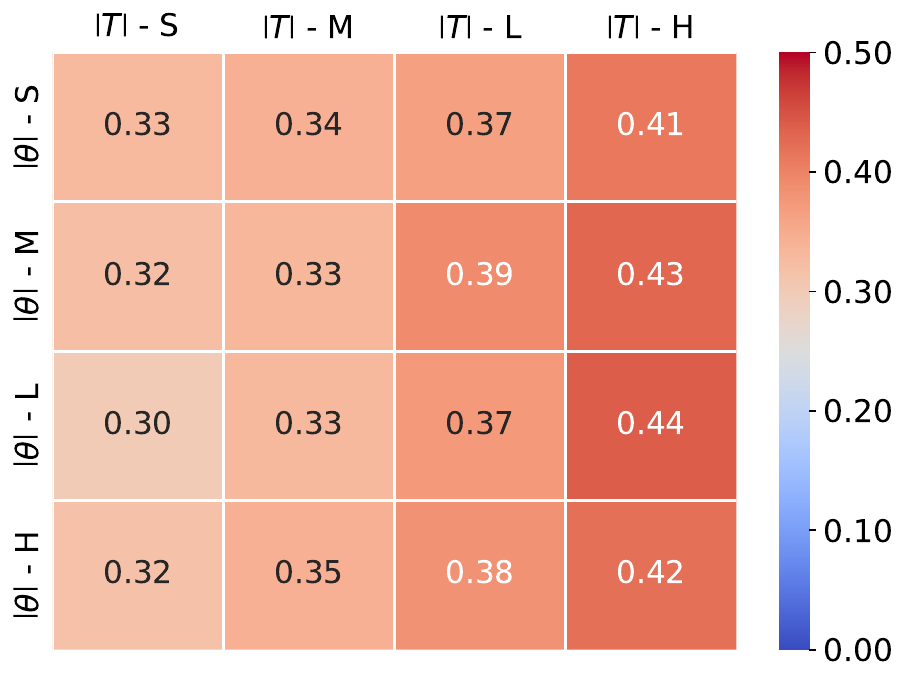}%
	\vspace{-1mm}}
	\hfil
	\subfloat[VeriFinger \cite{VeriFinger} ]{\includegraphics[width=.23\linewidth]{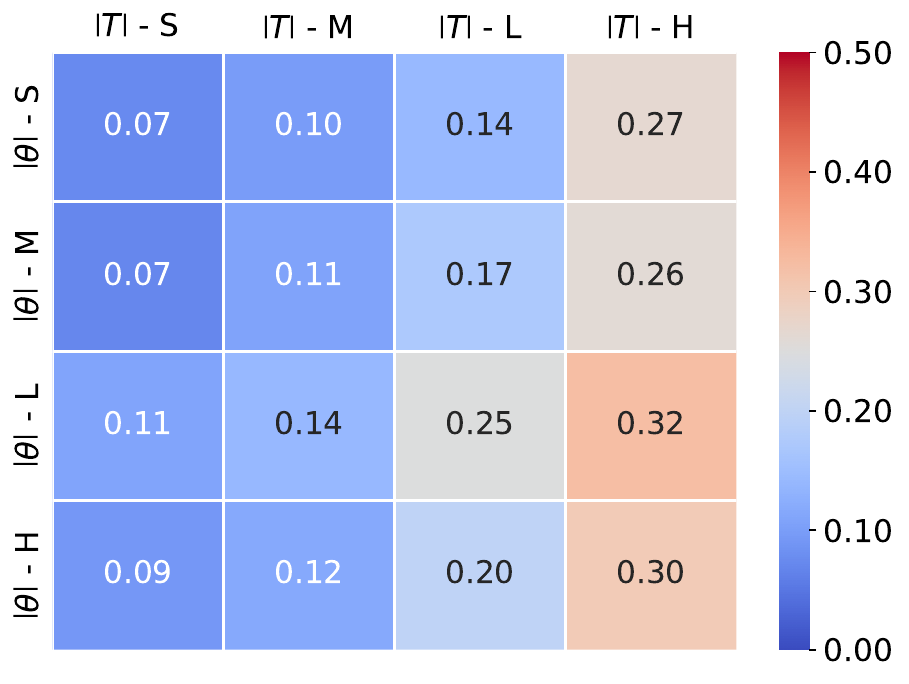}%
	\vspace{-1mm}}
	\hfil
	\subfloat[DeepPrint* \cite{engelsma2021learning} ]{\includegraphics[width=.23\linewidth]{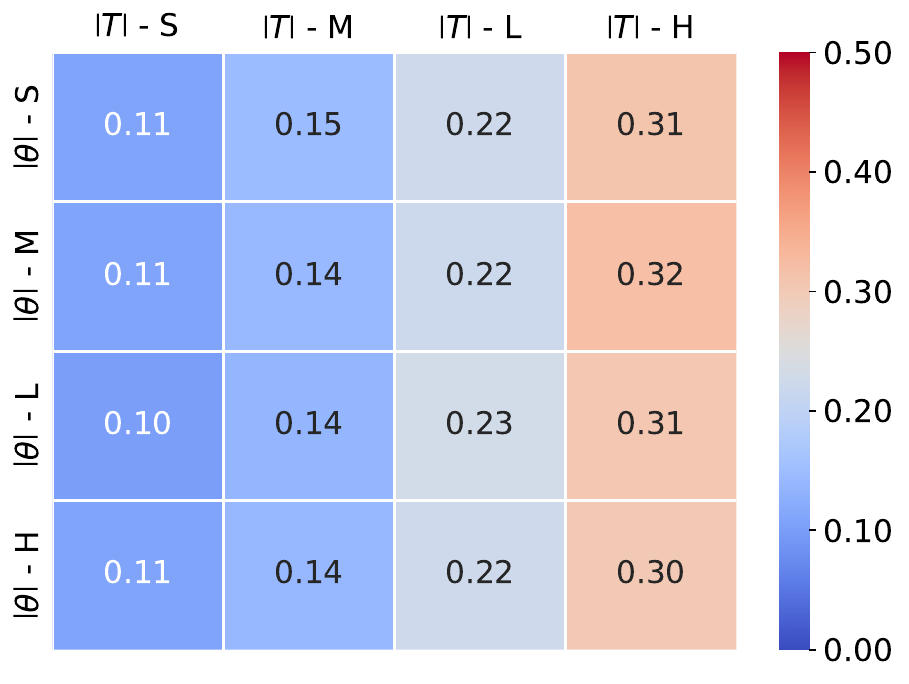}%
	\vspace{-1mm}}
	
	\par\vspace{1.5mm}
	
	\subfloat[DesNet* \cite{gu2021latent} ]{\includegraphics[width=.23\linewidth]{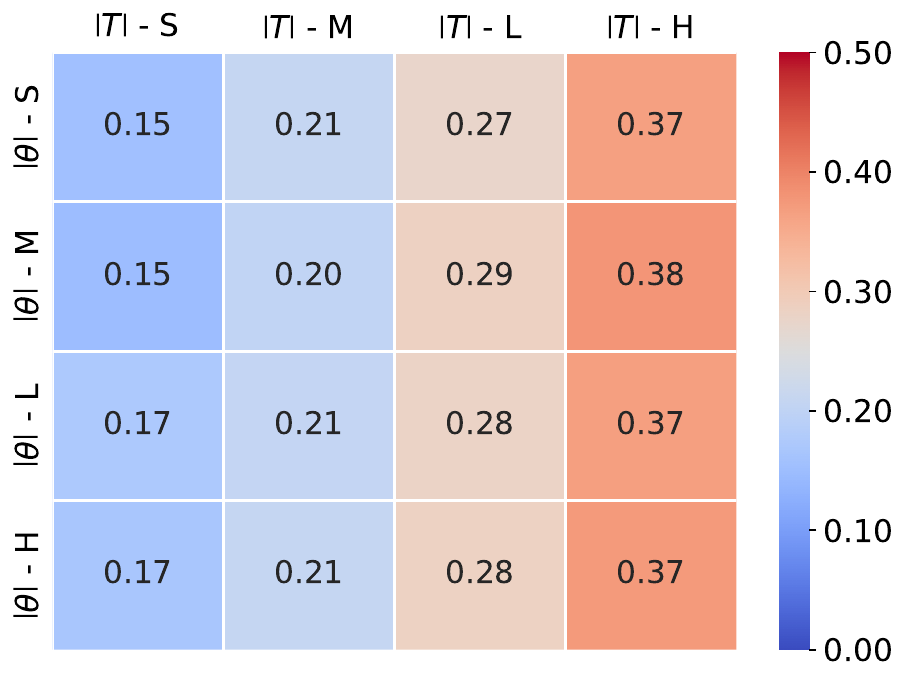}%
	\vspace{-1mm}}
	\hfil
	\subfloat[PFVNet \cite{he2022pfvnet}]{\includegraphics[width=.23\linewidth]{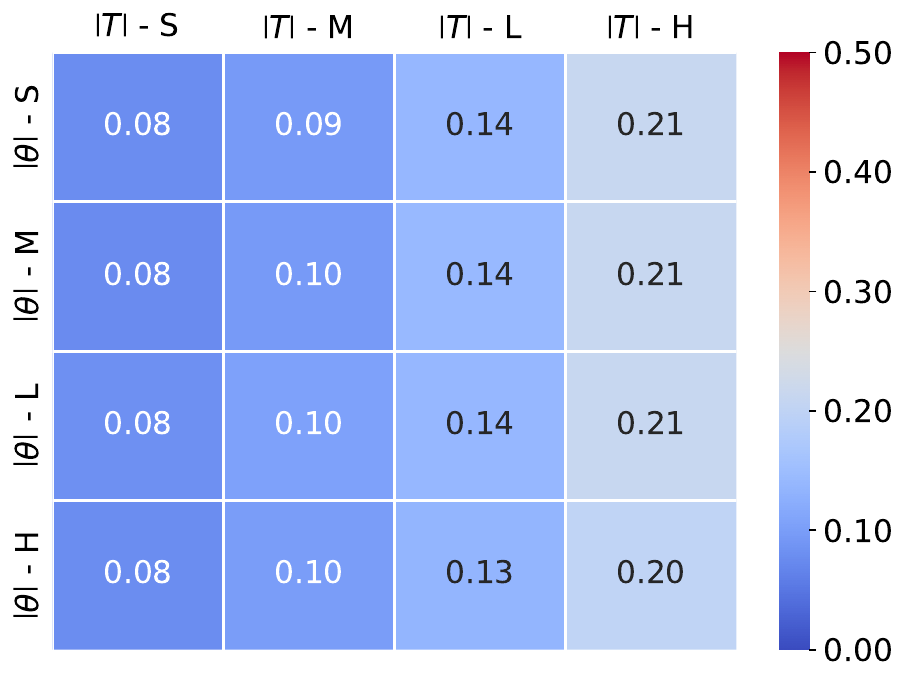}%
	\vspace{-1mm}}
	\hfil
	\subfloat[Proposed]{\includegraphics[width=.23\linewidth]{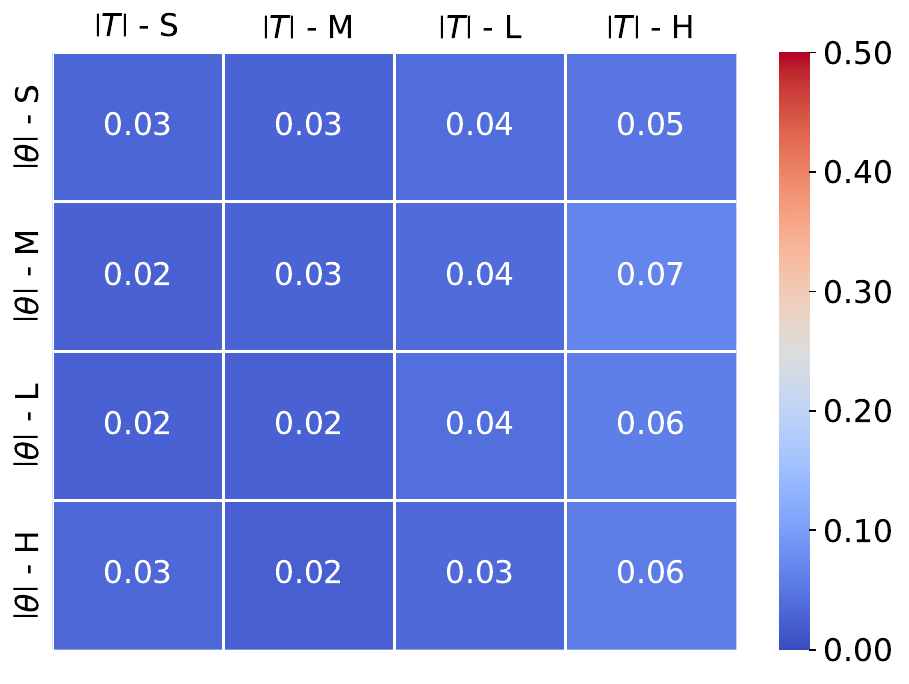}%
	\vspace{-1mm}}

	\caption{Equal Error Rate matrix under different degrees of relative rigid transformation on Hybrid DB\_B with image size $160 \times 160$. Genuine matches are divided into 16 subsets based on the value of relative rotation $\theta$ and translation $T$, which in terms of Small, Medium, Large, and Huge segmented evenly by numerical range. Impostor matches are not modified again.}
	\label{fig:matrix}
\end{figure*}

\begin{figure*}[h]
	\centering
	\subfloat[Hybrid DB\_B]{\includegraphics[width=.45\linewidth]{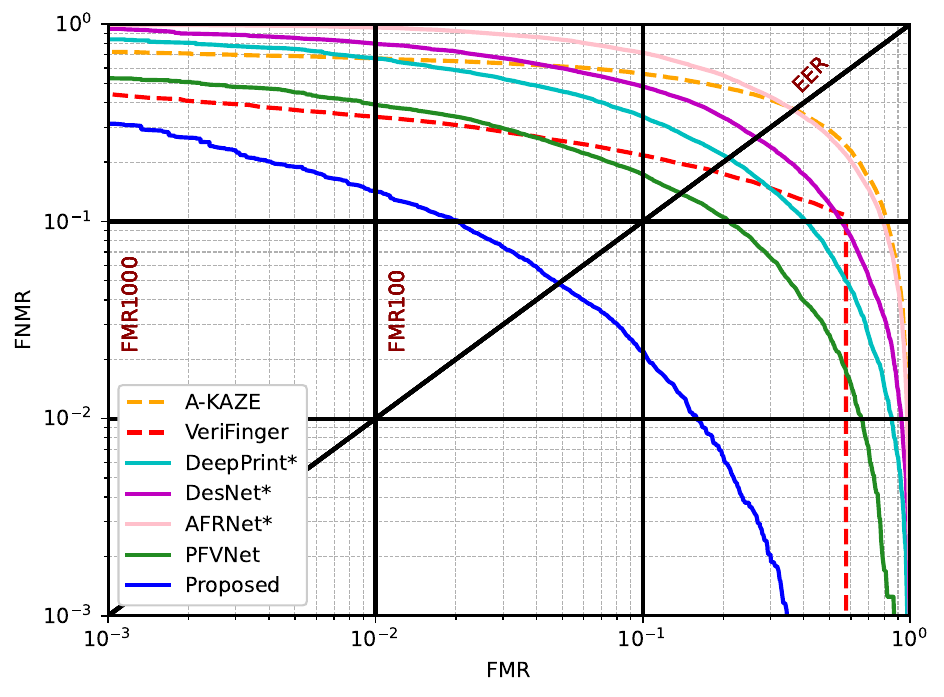}%
		\vspace{-1.5mm}}
	\hfil
	\subfloat[FVC2002 DB1\_A]{\includegraphics[width=.45\linewidth]{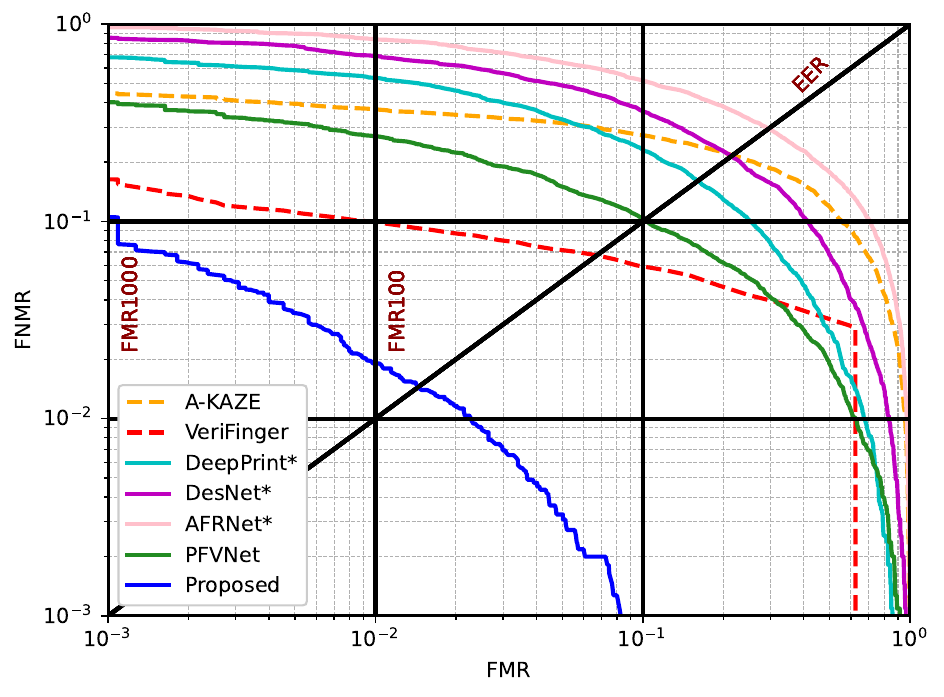}%
		\vspace{-1.5mm}}
	
	\par\vspace{1mm}
	
	\subfloat[FVC2002 DB3\_A]{\includegraphics[width=.45\linewidth]{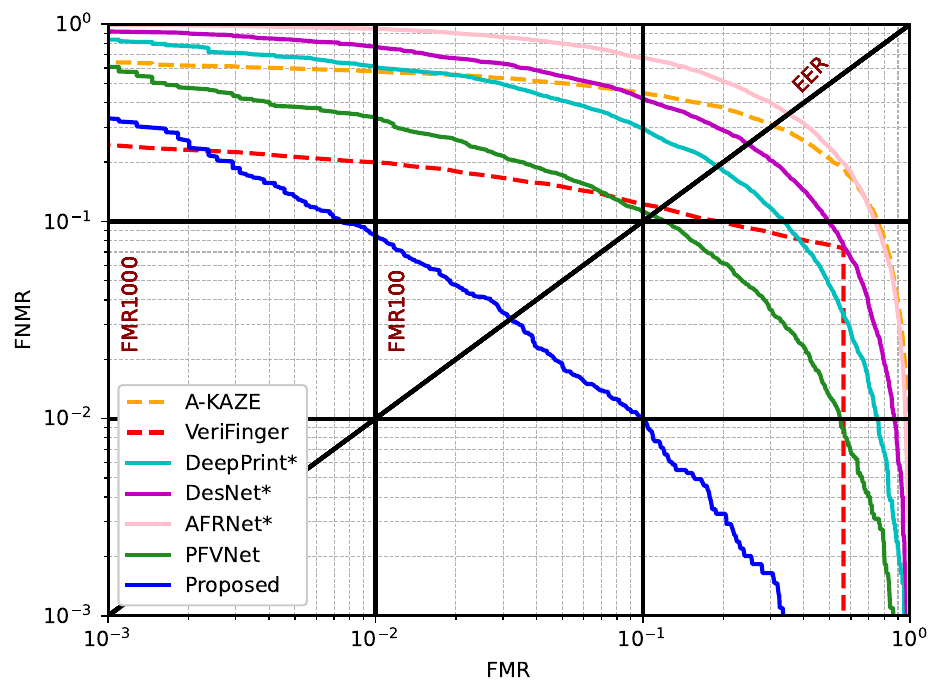}%
		\vspace{-1.5mm}}
	\hfil
	\subfloat[THU Small]{\includegraphics[width=.45\linewidth]{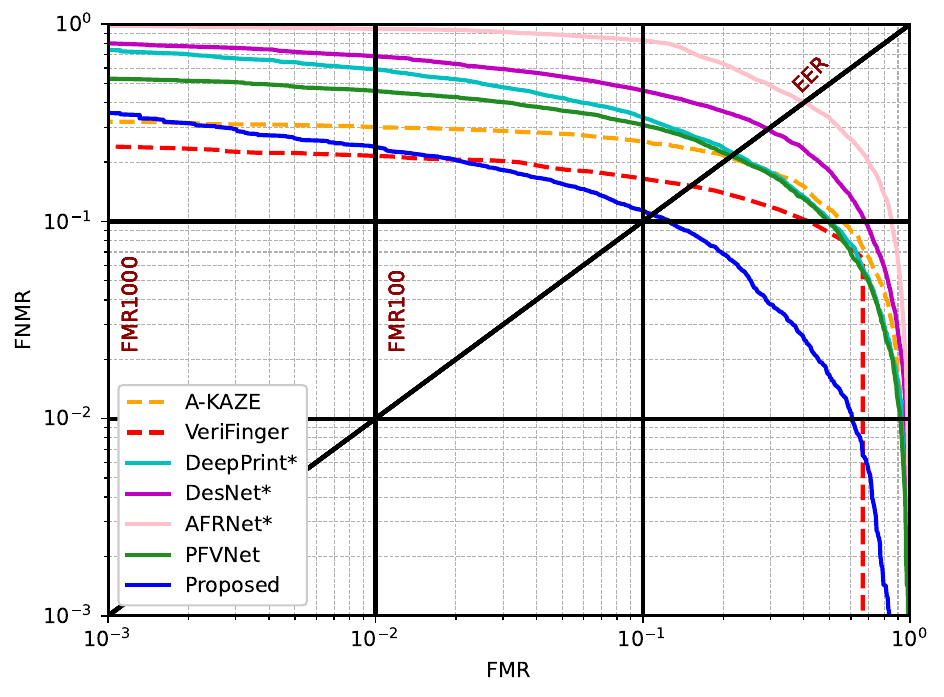}%
		\vspace{-1.5mm}}
	
	\caption{
	{
	DET curves of state-of-the-art algorithms and our approach on partial fingerprints of $160 \times 160$. Solid and dotted lines represent deep learning methods and traditional methods respectively.}}
	\label{fig:det}
\end{figure*}

In this subsection, four classic indicators are adopted to comprehensively evaluate methods from the perspective of different deployment scenarios, including:
\begin{itemize}
	\item Accuracy (ACC) under optimal thresholds, to intuitively evaluate the correctness of model prediction;
	\item Area Under Curve (AUC) based on Receiver Operating Characteristic (ROC) curves, to approximately evaluate the overall classification performance;
	\item True Accept Rate (TAR)  under specified False Accept Rate (FAR) of 1$\mathrm{e}$-3, which is often used to measure the security performance of biometric solutions.
	\item Equal Error Rate (EER), where both false alarm and impostor pass errors are equal.
\end{itemize}
Table \ref{tab:ex_matching_performance} presents the corresponding matching performance of state-of-the-art algorithms and our proposed approach on several representative datasets of fingerprints with different sizes.
Furthermore, we group the genuine matches on \emph{Hybrid DB\_B} according to their relative rigid transformation relationship and compare these algorithms under different rotations and translations using EER.
Fig. \ref{fig:matrix} shows the corresponding comparisons of representative methods for each type of matching framework.
Additionally, the Detection Error Tradeoff (DET) curves on all test databases are calculated in Fig. \ref{fig:det} to provide more complete information for analysis.

These encouraging experimental results strongly demonstrates the superiority of our proposed JIPNet, which outperforms previous advanced methods in almost all cases.
PFVNet \cite{he2022pfvnet} also shows remarkable performance, especially in indicators of ACC and AUC. 
However, it can be observed that the relative performance deteriorates markedly when training and testing data cross domains.
{
Region representation based methods\cite{engelsma2021learning,gu2021latent,grosz2024afr} are inferior to others, which is reasonable because the highly generalized form of fixed-length descriptors sacrifices many potential details for faster search speed.
On the other hand, emphasizing certain texture features located in non overlapping areas may also lead to incorrect mismatches.
That is to say, genuine matches with very small overlapping areas and impostor matches with very similar local patterns lack sufficient stable discrimination under such regional representations, which is also reflected in Fig. \ref{fig:ex_score_distribution}.}
Specifically, although localized descriptors with deeper dimensions are designed in DesNet \cite{gu2021latent}, their effectiveness is limited, as stated in Section \ref{subsec:score_distribution}.
On the other hand, the performance of all algorithms drops significantly as the image size decreases or distance increases, but our scheme still leads and exhibits attractive robustness and stability.
It is worth noting that the impact of rotation is not evident in Fig. \ref{fig:matrix}, because key points based matching score do not require pose rectification and the prediction error in other algorithms remains at a similar level across different relative angles (confirmed in the following experiments).

\begin{table*}[!ht]
	\renewcommand\arraystretch{1.5}
	\belowrulesep=-0.2pt
	\aboverulesep=-0.2pt
	\caption{Matching Performance of Different Score Level Fusion Strategies with image size $160 \times 160$.}
	\label{tab:ex_fusion}
	\vspace{-0.4cm}
	\begin{center}
		\begin{threeparttable}
				\tabcolsep=7pt
				\begin{tabular}{c | l | ll | lll}
					\toprule
					\multirow{2}{*}[-.5mm]{\textbf{Dataset}}  & \; \textbf{VeriFinger}\cite{VeriFinger} & \quad \textbf{PFVNet}\cite{he2022pfvnet} & \quad \textbf{PFVNet}\cite{he2022pfvnet} & \quad \textbf{Proposed} & \quad \textbf{Proposed} & \; \textbf{Proposed} \\
					{} & + DeepPrint*\cite{engelsma2021learning} & + VeriFinger\cite{VeriFinger} & + DeepPrint*\cite{engelsma2021learning} &  + VeriFinger\cite{VeriFinger} & + DeepPrint*\cite{engelsma2021learning} & + PFVNet\cite{he2022pfvnet}
					\\
					\midrule
					Hybrid DB\_B & 
					\qquad 0.57 {\scriptsize\textcolor{olive}{$\uparrow \! 0.01$}} & 
					\qquad 0.66 {\scriptsize\textcolor{olive}{$\uparrow \! 0.19$}} & 
					\qquad 0.47 {\scriptsize\textcolor{olive}{$-$}}& 
					\qquad \textbf{0.76} {\scriptsize\textcolor{olive}{$\uparrow \! 0.05$}} & 
					\qquad 0.74 {\scriptsize\textcolor{olive}{$\uparrow \! 0.03$}} & 
					\qquad 0.71 {\scriptsize\textcolor{olive}{$-$}}\\
					FVC2002 DB1\_A & 
					\qquad 0.86 {\scriptsize\textcolor{olive}{$\uparrow \! 0.01$}} & 
					\qquad 0.87 {\scriptsize\textcolor{olive}{$\uparrow \! 0.26$}} & 
					\qquad 0.62 {\scriptsize\textcolor{olive}{$\uparrow \! 0.01$}} & 
					\qquad 0.94 {\scriptsize\textcolor{olive}{$\uparrow \! 0.01$}}& 
					\qquad 0.92 {\scriptsize\textcolor{red}{$\downarrow \! 0.01$}}& 
					\qquad \textbf{0.95} {\scriptsize\textcolor{olive}{$\uparrow \! 0.02$}}\\
					FVC2002 DB3\_A & 
					\qquad 0.76 {\scriptsize\textcolor{olive}{$-$}} & 
					\qquad 0.77 {\scriptsize\textcolor{olive}{$\uparrow \! 0.38$}} & 
					\qquad 0.49 {\scriptsize\textcolor{olive}{$\uparrow \! 0.10$}} & 
					\qquad \textbf{0.87} {\scriptsize\textcolor{olive}{$\uparrow \! 0.20$}}& 
					\qquad 0.68 {\scriptsize\textcolor{olive}{$\uparrow \! 0.01$}}& 
					\qquad 0.73 {\scriptsize\textcolor{olive}{$\uparrow \! 0.06$}}\\
					THU Small & 
					\qquad 0.76 {\scriptsize\textcolor{olive}{$-$}} & 
					\qquad 0.77 {\scriptsize\textcolor{olive}{$\uparrow \! 0.30$}} & 
					\qquad 0.47 {\scriptsize\textcolor{olive}{$-$}} & 
					\qquad \textbf{0.80} {\scriptsize\textcolor{olive}{$\uparrow \! 0.10$}}& 
					\qquad 0.66 {\scriptsize\textcolor{red}{$\downarrow \! 0.04$}}& 
					\qquad 0.66 {\scriptsize\textcolor{red}{$\downarrow \! 0.04$}}\\
					\bottomrule
					
				\end{tabular}
			\begin{tablenotes}
				\item[] TAR @ FAR = 1$\mathrm{e}$-3 is reported. 
				\item[] Scores of bold method are used as the baselines of corresponding columns. `+' indicates another type of method for fusion.
			\end{tablenotes}
		\end{threeparttable}
	\end{center}
\end{table*}

As analyzed above, these compared methods focus on different aspects of fingerprint attributes, exhibiting differentiated trends in score distribution and matching advantages.
This phenomenon motivates us to conduct multiple fusion strategies to further observe their complementarity.
Specifically, scores from two sets are linearly weighted and summed with the best classification accuracy.
As represented in Table \ref{tab:ex_fusion}, most fusions effectively improves the corresponding performance, especially the introduction of minutiae information (VeriFinger) into other methods that rely on texture features, while our method still leads the way.

\subsection{Alignment Accuracy}

\begin{figure}[!t]
	\centering
	\subfloat[]{\includegraphics[width=.5\linewidth]{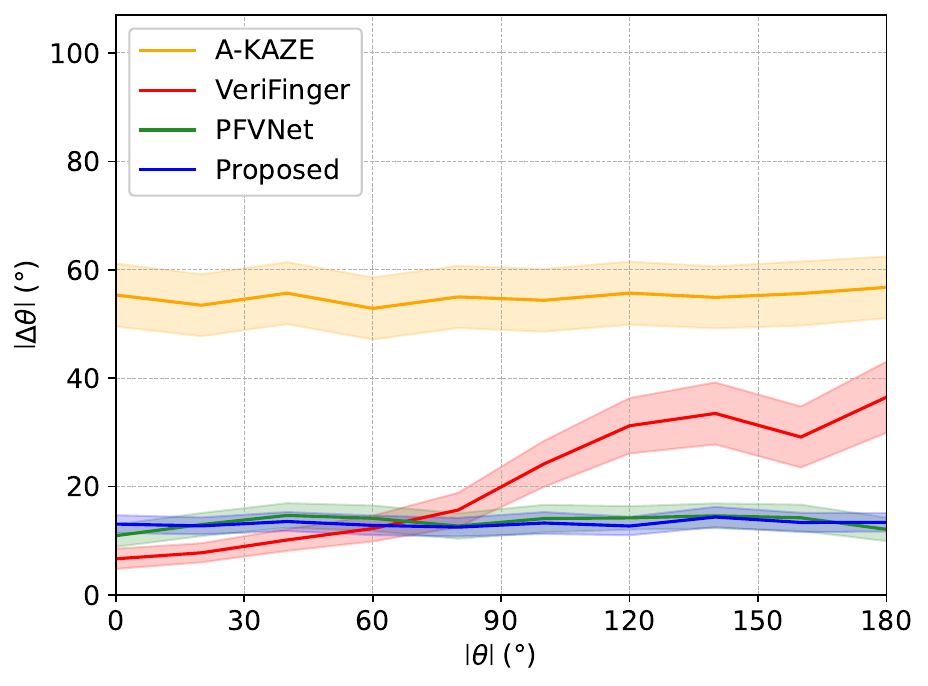}%
		\vspace{-1.5mm}}
	\hfil
	\subfloat[]{\includegraphics[width=.5\linewidth]{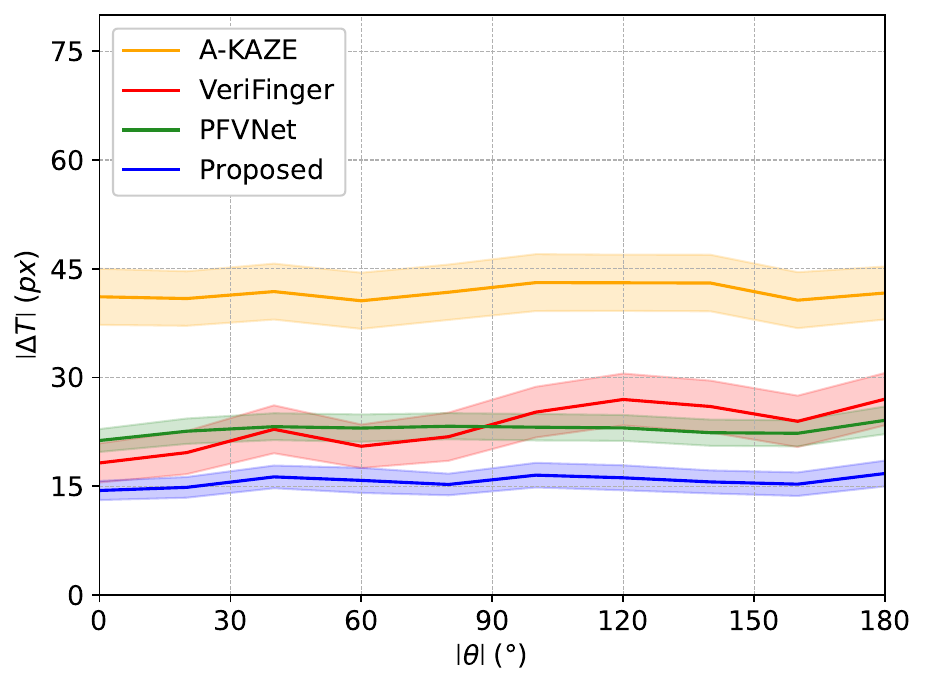}%
		\vspace{-1.5mm}}
	
	\subfloat[]{\includegraphics[width=.5\linewidth]{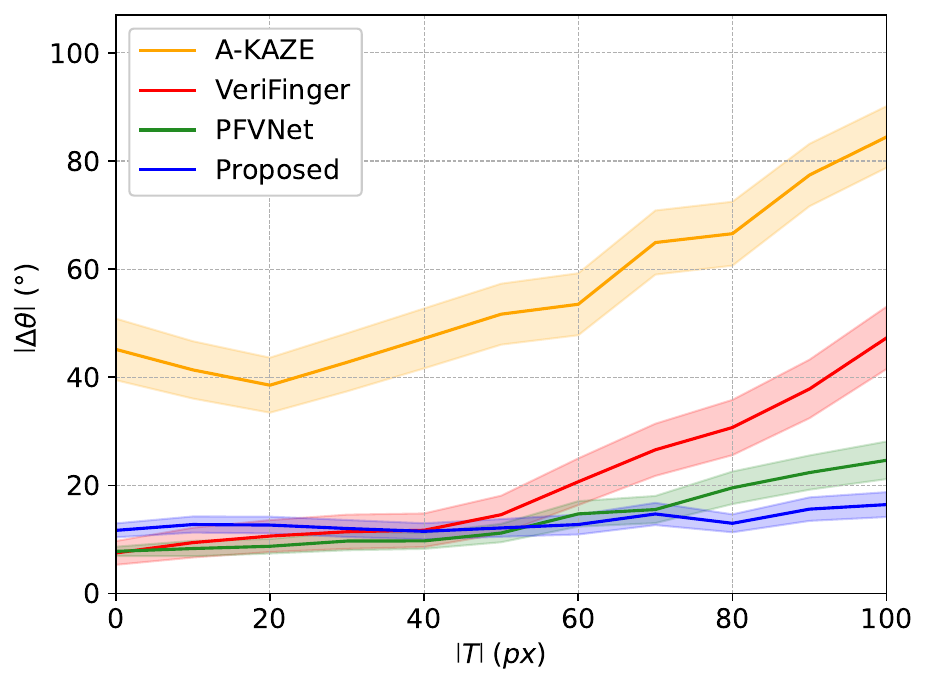}%
		\vspace{-1.5mm}}
	\hfil
	\subfloat[]{\includegraphics[width=.5\linewidth]{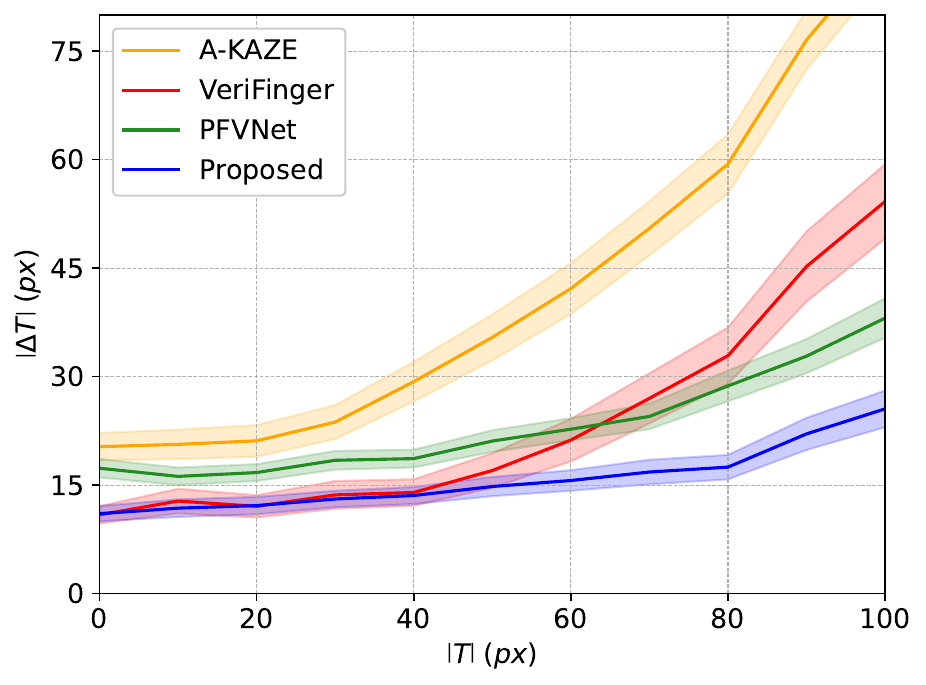}%
		\vspace{-1.5mm}}

	\caption{Estimation errors of alignment on Hybrid DB\_B in image size $160 \times 160$. (a) Rotation errors and (b) Translation errors under varying $\left|\theta\right|$. (c) Rotation errors and (d) Translation errors under varying $\left|T\right|$. Lines and corresponding light-colored bands reflect the mean and standard deviation.}
	\label{fig:align}
\end{figure}

The original intention of introducing the pose estimation sub stage in our network is to help it better understand the spatial contrast relationship and thus facilitate partial fingerprint verification.
Nevertheless, we still assess the alignment accuracy as it is indeed a specific aspect that can be effectively compared, allowing us to gauge their capability to rectify relative positions and potential influence on subsequent matching.
Similar to \cite{he2022pfvnet}, we express the estimation error in the form of translation and rotation as:
\begin{equation}
	\begin{aligned}
		|\Delta \theta| &=|\theta^* - \theta| \; , \\
		|\Delta T| &= \sqrt{(t_\mathrm{x}^* - t_\mathrm{x})^2+(t_\mathrm{y}^* - t_\mathrm{y})^2} \; , \\
	\end{aligned}
	\label{eq:delta_align}
\end{equation}
where symbols are defined the same as in Equation \ref{eq:loss_reg}.
As shown in Fig. \ref{fig:align}, most methods can provide effective predictions within an acceptable range, while our approach has a slight edge.
Besides, it can be seen that estimation errors of both rotation and translation increase significantly with distance and remaining roughly stable for angle.
One convincing explanation is that distance greatly alters the overlap area, which drastically affects the internal feature comparison and subsequent decision-making process.

\subsection{Ablation Study}

Extensive ablation experiments are conducted to examine the effects of different modules and strategies introduced in our network.
{
	EER and $\mathcal{L}_{\mathrm{reg}}$ (defined in Equation \ref{eq:loss_reg}) are used to measure the performance of matching and alignment, respectively.
}
As shown in Table \ref{tab:ex_ablation}, overall, the CNN based approach outperforms Vision transformer (ViT) based approach, while the CNN-Transformer hybrid series achieves the best performance.
{
We attribute it to the respective advantages of convolutional and transformer blocks in extracting local features and establishing global connections.
In addition, the introduction of parallel relative alignment estimation head achieves great success in EER metric across all three types of network architectures, which proves that there is indeed an exploitable promotion relationship between pose prediction and identity verification.
Furthermore, the joint estimation strategy also shows certain positive effect on pose prediction task when applied in networks containing ViT decoders.
At the same time, the additional alignment head can be pruned during deployment so that the matching speed will not be affected.
Finally, the comparison between last two rows demonstrated that our proposed lightweight pre-training task can further improve the performance at a small extra training time cost.
A convincing explanation is that it ensures the extracted features contain complete pattern information of fingerprint ridges, while also performing a certain degree of denoising.
In this way, the inputs for subsequent two modules, namely identity verification and pose prediction, is transferred from the image level to the feature level, which assists the network to  better focus on key features and perform joint optimization more seamlessly.}

\begin{table}[!t]
	\renewcommand\arraystretch{1.5}
	\belowrulesep=-0.2pt
	\aboverulesep=-0.2pt
	\caption{{
			Ablation Study of the Proposed Network with Different Modules and Strategies on {FVC2002 DB1\_A} with image size $160\times160$.}}
	\label{tab:ex_ablation}
	\vspace{-0.4cm}
	\begin{center}
		\begin{threeparttable}
			\tabcolsep=6.5pt
			\begin{tabular}{cccccc|cc}
				\toprule
				\multicolumn{2}{c}{{\textbf{Encoder}}\tnote{\,*}} &
				\multicolumn{2}{c}{{\textbf{Decoder}}\tnote{\,*}} &
				\multirow{2}{*}[-.5mm]{\makecell[c]{\textbf{+ ID}\\ \textbf{Head}\tnote{\,*}}} &
				\multirow{2}{*}[-.5mm]{\makecell[c]{\textbf{+ Pose}\\ \textbf{Head}\tnote{\,*}}} &
				\multirow{2}{*}[-.5mm]{{\textbf{EER}}} &
				\multirow{2}{*}[-.5mm]{{$\bm{\mathcal{L}_{\mathrm{reg}}}$}}\\
				\cmidrule(lr){1-2}\cmidrule(lr){3-4}
				\scriptsize\textbf{CNN} & \scriptsize\textbf{ViT} &
				\scriptsize\textbf{CNN} & \scriptsize\textbf{ViT} &
				{} & {} & {} & {}\\
				\midrule
				{\checkmark} & {} & {\checkmark} & {} & {\checkmark} & {} & 0.25  & {-}\\
				{\checkmark} & {} & {\checkmark} & {} & {} & {\checkmark} & {-}  & 18.2\\
				{\checkmark} & {} & {\checkmark} & {} & {\checkmark}& {\checkmark} & 0.21  & {20.1}\\
				\midrule
				{} & {\checkmark} & {} & {\checkmark} & {\checkmark} & {} & 0.47 & {-}\\
				{} & {\checkmark} & {} & {\checkmark} & {} & {\checkmark} & {-} & 27.2\\
				{} & {\checkmark} & {} & {\checkmark} & {\checkmark} & {\checkmark} & 0.29 & {26.7}\\
				\midrule
				{\checkmark} & {} & {} & {\checkmark} & {\checkmark} & {} & 0.17 & {-}\\
				{\checkmark} & {} & {} & {\checkmark} & {} & {\checkmark} & {-} & 10.6\\
				{\checkmark} & {} & {} & {\checkmark} & {\checkmark} & {\checkmark} & 0.07 & {9.2}\\
				\midrule
				\rowcolor{black!10}
				{\checkmark\tnote{$\S$}} & {} & {} & {\checkmark} & {\checkmark} & {\checkmark} & \textbf{0.01} & \textbf	{1.1} \\
				\bottomrule
			\end{tabular}
			\begin{tablenotes}
				\item[$\S$] {
					indicates the corresponding module has been pretrained through the process shown in Figure \ref{fig:network_pretrain}.}
				\item[*] each represents specific stages in proposed network, specifically \emph{Encoder} for Stage 1 \& 2, \emph{Decoder} for Stage 3, \emph{ID Head} for Stage 4\,-\,C, \emph{Pose Head} for Stage 4\,-\,R.
			\end{tablenotes}
		\end{threeparttable}
	\end{center}
\end{table}

\subsection{Visual Analysis}

\begin{figure*}[!ht]
	\centering	
	\includegraphics[width=.9\linewidth]{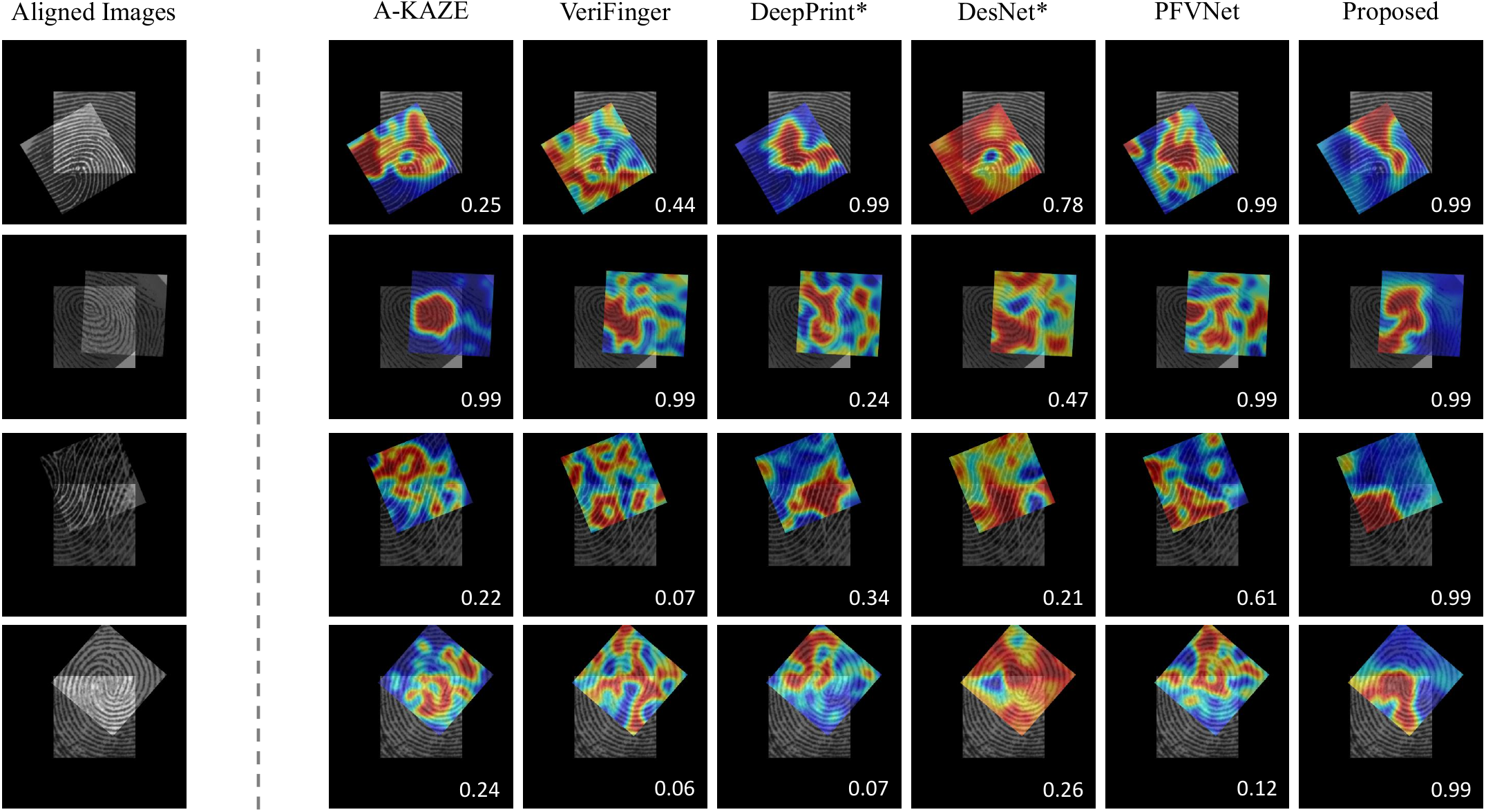}
\vspace*{-1mm}
\caption{Representative visualization results of genuine matches based on occlusion sensitivity \cite{chen2022query2set}. 
	Red areas indicate a large influence on matching judgment, while blue areas indicate the opposite. 
	Numbers on the bottom right are the accuracy of corresponding methods when they are verified.}
\label{fig:example_good}
\end{figure*}

\begin{figure}[!t]
	\centering	
	\includegraphics[width=.87\linewidth]{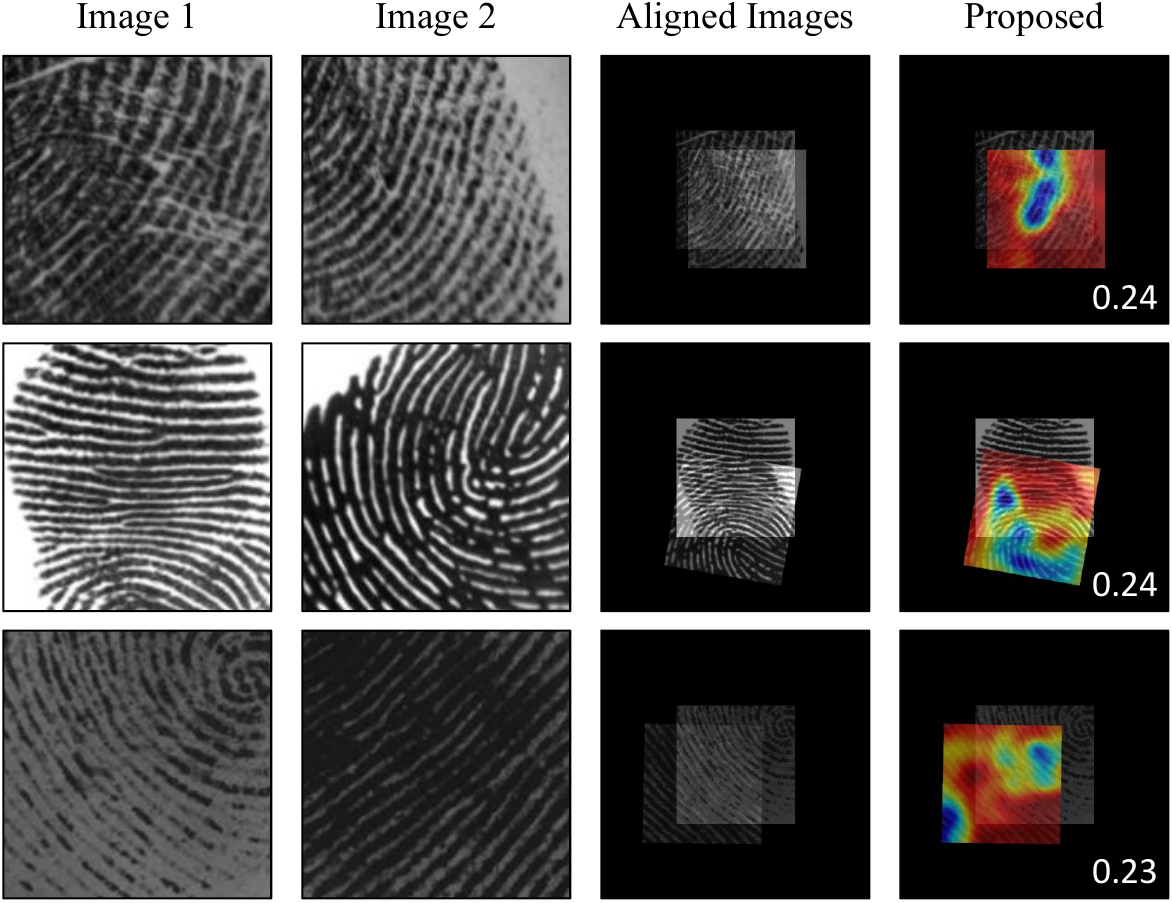}
\vspace*{-2mm}
\caption{Challenging cases where our approach fails. The way of visualization is the same as Fig. \ref{fig:example_good}.}
\label{fig:example_bad}
\end{figure}

In order to provide more specific and intuitive comparison with interpretability, we employ occlusion sensitivity \cite{chen2022query2set} to approximate the contribution of distinct local areas to the overall matching judgment.
Fig. \ref{fig:example_good} shows four representative visualization results.
Query fingerprint in images pairs is centered and search fingerprint is rigidly aligned based on the ground truth as background.
Subsequently, a $32\times32$ mask is slid and applied to search fingerprints and the differences between scores before and after occlusion are calculated, which are normalized and overlaid as the heatmap of search fingerprint.
Considering that there is no reference relationship between the matching scores directly obtained (or mapped) by different algorithms, we convert these scores into a comparable probability form and attached them in corresponding visualization results.
Specifically, given the overall distributions (shown in Fig. \ref{fig:ex_score_distribution}), the verification accuracy can be expressed as:
\begin{equation}
	\begin{aligned}
		P\left(y_1|s\right) &= \frac{P\left(s|y_1\right)P\left(y_1\right) }{P\left(s|y_1\right)P\left(y_1\right)+P\left(s|y_0\right)P\left(y_0\right)} \\
		&= \frac{f_{\mathrm{1}}\left(s\right)P\left(y_1\right)}{f_{\mathrm{1}}\left(s\right)P\left(y_1\right)+f_{\mathrm{0}}\left(s\right)P\left(y_0\right)} \;,\\
	\end{aligned}
	\label{eq:visual_map}
\end{equation}
where $s$ is the genuine matching score, $P$ and $f$ represent the probability and probability density, $y$ represents the identity to be verified (1 is the same, 0 is different).
It can be seen that algorithms based on key points \cite{mathur2016methodology,VeriFinger} and region representation \cite{engelsma2021learning,gu2021latent} expose their respective shortcomings  when there is a lack of sufficient features in overlapping areas (row 1) and significant differences in ridge texture of non overlapping areas (row 2).
PFVNet \cite{he2022pfvnet} has clear advantages in both cases, even with minimal overlap (row 3). 
However, it fails in the last example with confusion in rotation.
At the same time, our method demonstrates optimal stability and robustness in focusing on overlapping regions and gives correct decisions with the highest accuracy.

Meanwhile, some failure cases in Fig. \ref{fig:example_bad} show that current JIPNet still needs to be improved in some extreme scenarios, such as: (i) sever image defacement caused by incorrect collection conditions (row 1); (ii) significant misalignment of ridges caused by finger distortion (row 2); (iii) extremely weak texture, which refers to almost no minutiae or changes in orientation, as well as large modal differences (row 3).

\subsection{Efficiency}

\begin{figure}[!t]
	\centering	
	\includegraphics[width=1\linewidth]{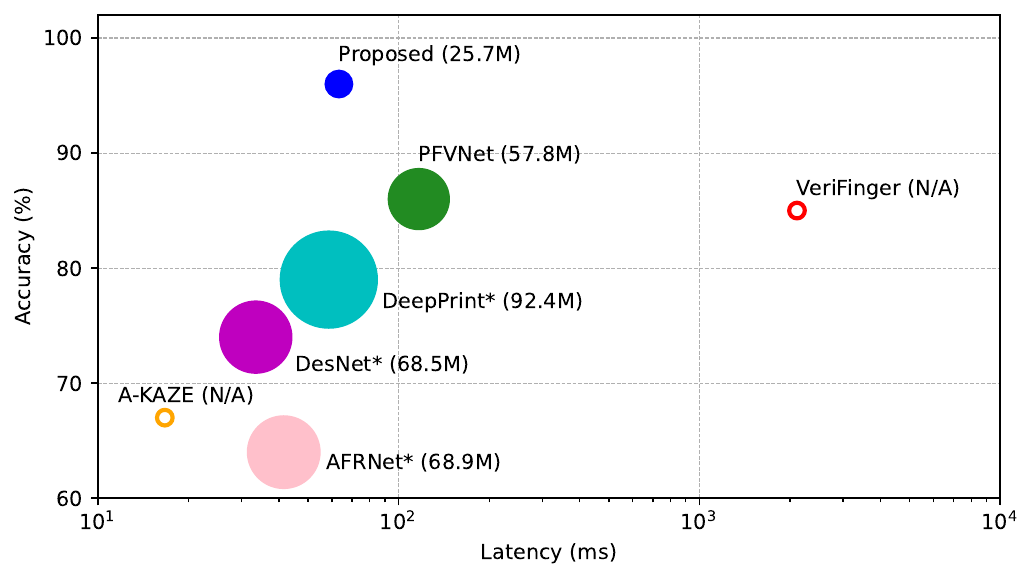}
\vspace*{-2mm}
\caption{{
		Comparisons of the trade-off between performance and efficiency on Hybrid DB\_B in image size $160 \times 160$. The size of solid points represents the model size of deep learning methods, while hollow points represent non learning algorithms.
}}
\label{fig:tradeoff}
\end{figure}

In this subsection, we present a comparison of deployment efficiency across different algorithms.
As shown in Fig \ref{fig:tradeoff}, our proposed method achieves a new state-of-the-art trade-off between matching performance, speed, and size.
This clearly highlights our streamlined joint estimation framework of pose and identity, as well as the appropriate and efficient network design.
The calculated time covers a complete process from inputting an original image pair to outputting the final estimation result.
All algorithms except for VeriFinger \cite{VeriFinger} (encapsulated SDK in C\,/\,C++) are implemented in Python and the batch size is set to 1, which are deployed on a single NVIDIA GeForce RTX 3090 GPU with an Intel Xeon E5-2680 v4 CPU @ 2.4 GHz.

\section{Conclusion}\label{sec:conclusion}
In this paper, we propose a joint framework of identity verification and pose alignment for partial fingerprints.
A novel CNN-Transformer hybrid network, named JIPNet, is presented to combine their advantages in feature extraction and information interaction, promoting the attention and utilization of valuable information.
In addition, a lightweight pre-training task is designed to improve the representation ability of feature encoder by reconstructing enhanced images.
Comprehensive experiments on multiple databases demonstrate the effectiveness and superiority of our proposed method.
Future studies will focus on further improving the performance in more challenging scenarios such as extreme low quality and small overlap, and extending our proposed method to one-to-many matching tasks.

{
	\bibliographystyle{IEEEtran}
	\bibliography{egbib}{}
}

\begin{IEEEbiography}[{\includegraphics[width=1in,height=1.25in,clip,keepaspectratio]{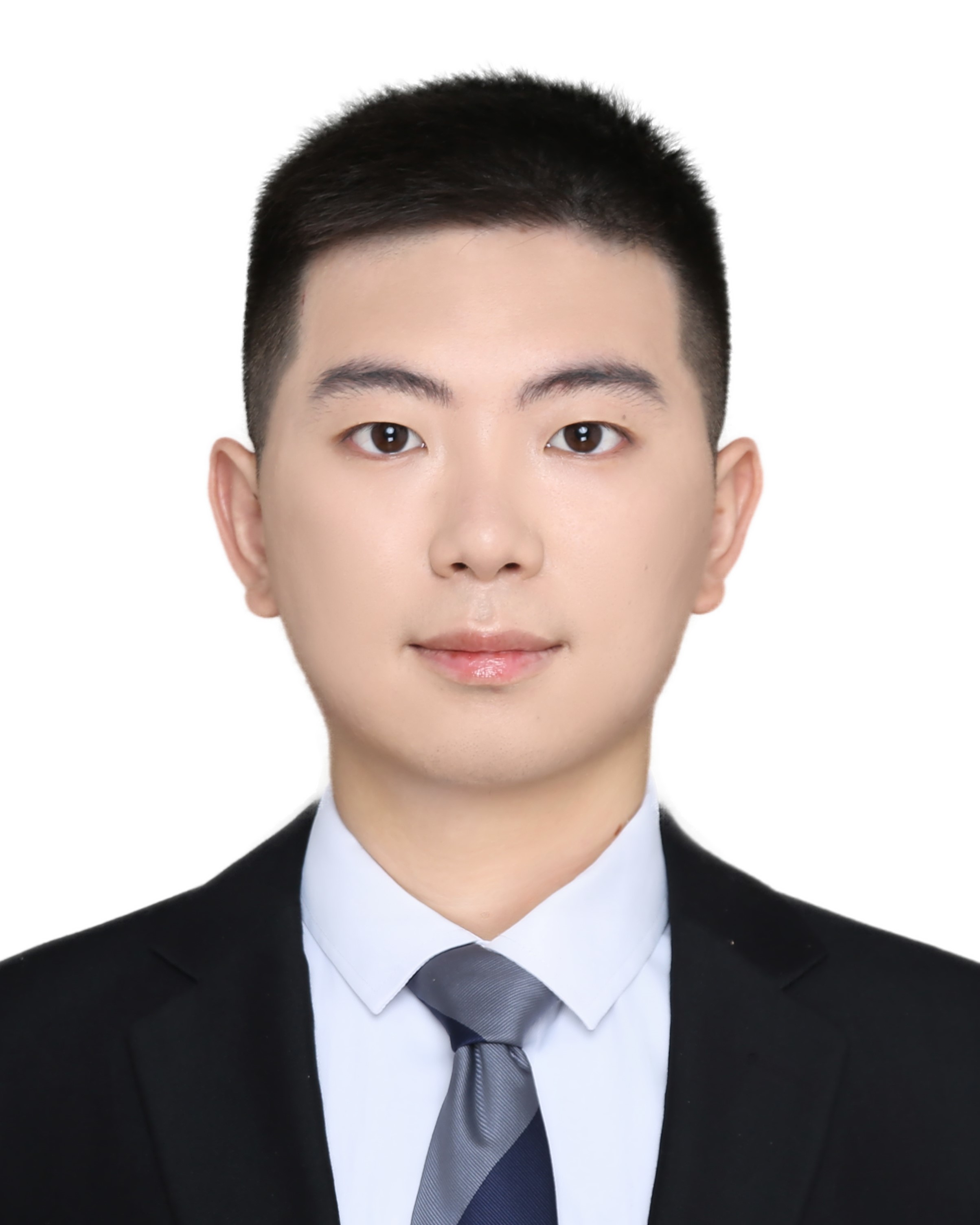}}]{Xiongjun Guan} received the B.S. degree from the Department of Automation, Tsinghua University, Beijing,
	China, in 2021, where he is currently pursuing the Ph.D. degree under the supervision of Prof.
	Jianjiang Feng with the Department of Automation. His research interests include fingerprint
	recognition, computer vision and pattern recognition.
\end{IEEEbiography}

\begin{IEEEbiography}[{\includegraphics[width=1in,height=1.25in,clip,keepaspectratio]{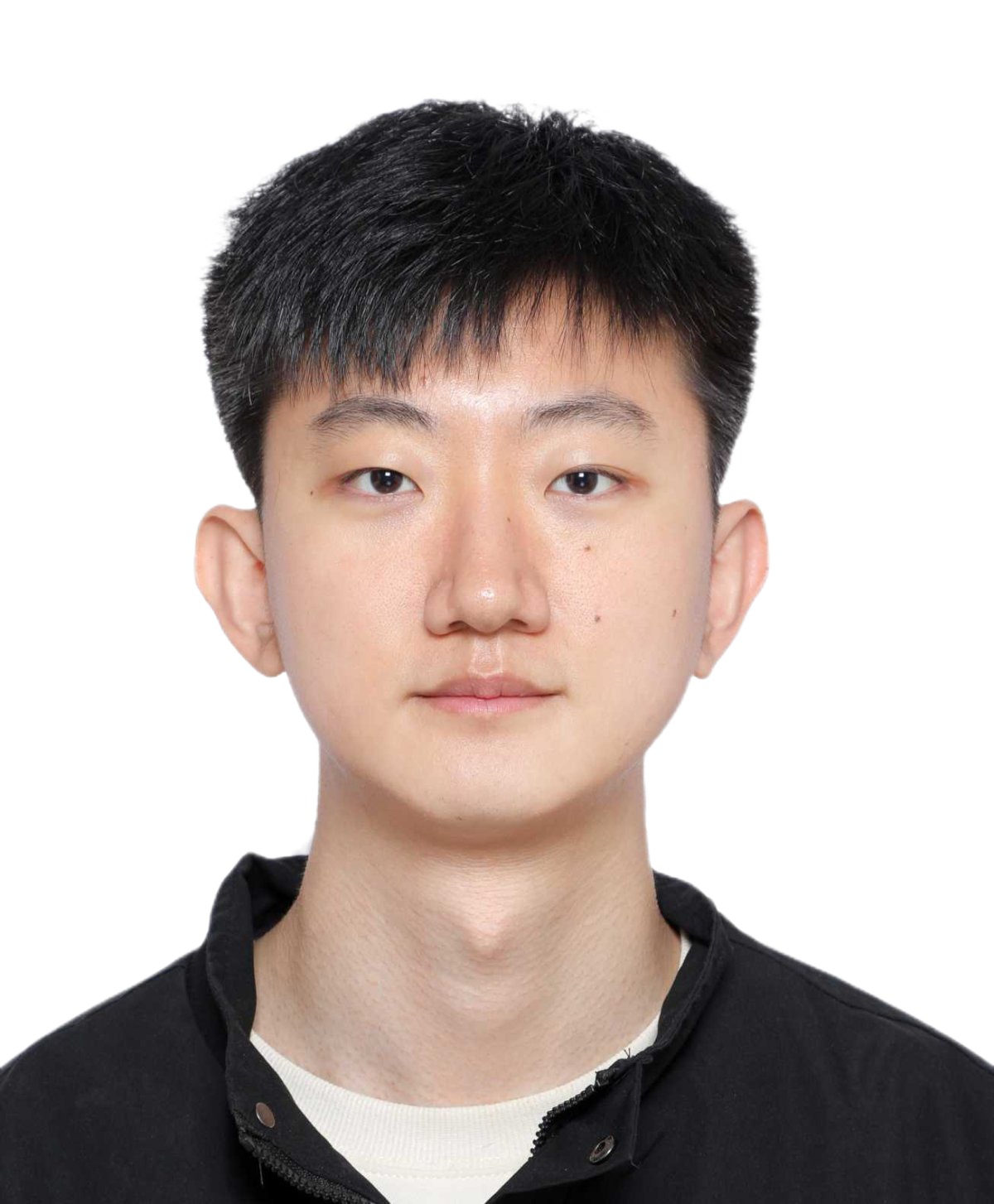}}]{Zhiyu Pan} received his Bachelor of Engineering (BEng) degree in Electronic Science and Technology from Beijing Institute of Technology, China, in 2020. He is currently pursuing a Ph.D. degree in the Department of Automation at Tsinghua University. His research interests include biometrics, human action analysis, and computer vision. Specifically, his current work focuses on fingerprint recognition, multi-modal learning, and related areas.
\end{IEEEbiography}

\begin{IEEEbiography}[{\includegraphics[width=1in,height=1.25in,clip,keepaspectratio]{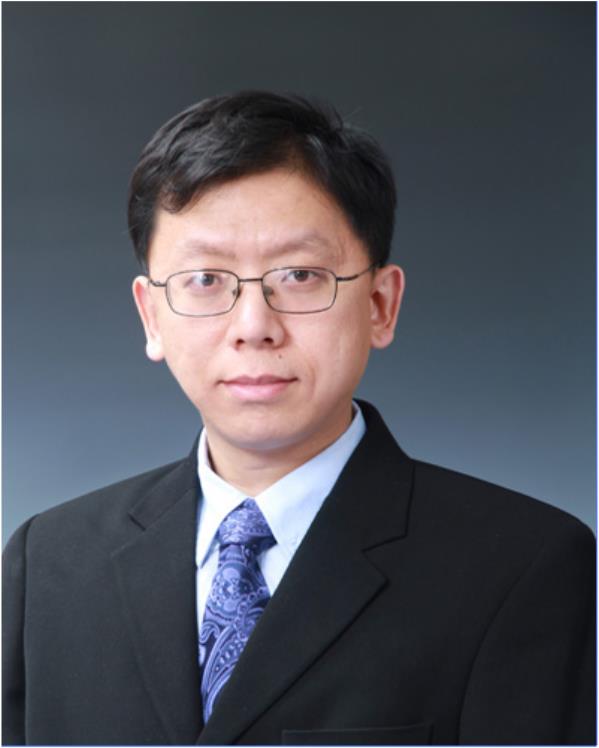}}]{Jianjiang
		Feng} received the B.Eng. and Ph.D. degrees from the School of Telecommunication Engineering,
	Beijing University of Posts and Telecommunications, China, in 2000 and 2007, respectively. From
	2008 to 2009, he was a Post-Doctoral Researcher with the PRIP Laboratory, Michigan State
	University. He is currently an Associate Professor with the Department of Automation, Tsinghua
	University, Beijing. His research interests include fingerprint recognition and computer vision.
\end{IEEEbiography}

\begin{IEEEbiography}[{\includegraphics[width=1in,height=1.25in,clip,keepaspectratio]{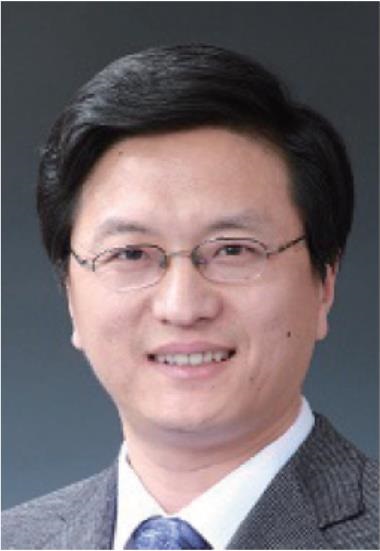}}]{Jie
		Zhou} received the B.S. and M.S. degrees from the Department of Mathematics, Nankai University,
	Tianjin, China, in 1990 and 1992, respectively, and the Ph.D. degree from the Institute of Pattern
	Recognition and Artificial Intelligence, Huazhong University of Science and Technology, Wuhan,
	China, in 1995. From 1995 to 1997, he served as a Post-Doctoral Fellow with the Department of
	Automation, Tsinghua University, Beijing, China. Since 2003, he has been a Full Professor with the
	Department of Automation, Tsinghua University. His research interests include computer vision,
	pattern recognition, and image processing. In recent years, he has authored more than 300 papers in
	peer-reviewed journals and conferences. Among them, more than 100 papers have been published in top
	journals and conferences such as the IEEE Transactions on Pattern Analysis and Machine
	Intelligence, IEEE Transactions on Image Processing, and CVPR. He is an associate editor for the
	IEEE Transactions on Pattern Analysis and Machine Intelligence and two other journals. He received
	the National Outstanding Youth Foundation of China Award. He is a Fellow of the IAPR and a senior
	member of the IEEE.
\end{IEEEbiography}

\end{document}